\documentclass{article}

\usepackage{microtype}
\usepackage{graphicx}
\usepackage{subcaption}
\usepackage{booktabs} % for professional tables

\usepackage[preprint]{ICML2026/icml2026}

\usepackage{amsmath}
\usepackage{amssymb}
\usepackage{mathtools}
\usepackage{amsthm}

\usepackage[algpseudocode=false]{signalpreamble}
\newacronym{FL}{FL}{federated learning}
\newacronym{PFL}{PFL}{personalized federated learning}
\newacronym{MLP}{MLP}{multi-layer perceptron}
\newacronym{HN}{HN}{hypernetwork}
\theoremstyle{plain}

\theoremstyle{definition}

\theoremstyle{remark}

\usepackage[textsize=tiny]{todonotes}

\icmltitlerunning{Representation-Aligned Multi-Scale Personalization for Federated Learning}

\begin{document}

\twocolumn[
  \icmltitle{Representation-Aligned Multi-Scale Personalization for Federated Learning}



  \icmlsetsymbol{equal}{*}

  \begin{icmlauthorlist}
    \icmlauthor{Wenfei Liang}{yyy}
    \icmlauthor{Wee Peng Tay}{yyy}
    % \icmlauthor{xxx}{yyy}
    
  \end{icmlauthorlist}

  \icmlaffiliation{yyy}{Department of Electrical and Electronic Engineering, Nanyang Technological University, Singapore}
  % \icmlaffiliation{yyy}{x}

  \icmlcorrespondingauthor{Wenfei Liang}{wenfei001@e.ntu.edu.sg}


  \vskip 0.3in
]



\printAffiliationsAndNotice{Code: \url{https://github.com/CarrieWFF/FRAMP}.}
% no special notice (required even if empty)

\begin{abstract}
In federated learning (FL), accommodating clients with diverse resource constraints remains a significant challenge. A widely adopted approach is to use a shared full-size model, from which each client extracts a submodel aligned with its computational budget. However, regardless of the specific scoring strategy, these methods rely on the same global backbone, limiting both structural diversity and representational adaptation across clients.
This paper presents FRAMP, a unified framework for personalized and resource-adaptive federated learning. Instead of relying on a fixed global model, FRAMP generates client-specific models from compact client descriptors, enabling fine-grained adaptation to both data characteristics and computational budgets. Each client trains a tailored lightweight submodel and aligns its learned representation with others to maintain global semantic consistency. Extensive experiments on vision and graph benchmarks demonstrate that FRAMP enhances generalization and adaptivity across a wide range of client settings.
\end{abstract}

%#############################
\section{Introduction}

\Gls{FL} has emerged as a promising paradigm for distributed model training, enabling multiple clients to collaboratively learn a shared model without exchanging raw data \cite{konevcny2016federated}. In this setting, a central server coordinates the training by aggregating locally updated models from clients. However, FL faces challenges in real-world deployments due to system heterogeneity, where clients differ significantly in computational capabilities \cite{diao2020heterofl,lin2020ensemble}.
A practical solution is submodel extraction, which enables model heterogeneity by allowing each client to train a sparse subnetwork derived from a shared full-size model.

Static submodel extraction \cite{diao2020heterofl,kim2022depthfl,ilhan2023scalefl,kang2023nefl,wang2023flexifed,yi2024fedp3} determines each client’s submodel architecture before training and keeps it fixed throughout the entire process. However, such static designs fail to account for the evolving model dynamics, often resulting in suboptimal performance.
In contrast, dynamic submodel extraction \cite{caldas2018expanding,horvath2021fjord,alam2022fedrolex,deng2022tailorfl,liao2023adaptive,chen2023efficient,wang2024feddse,wu2024fiarse} updates submodel structures across rounds to better track model evolution. For example, FedRolex \cite{alam2022fedrolex} employs a rolling strategy treating all parameters equally, while FIARSE \cite{wu2024fiarse} selects submodels based on parameter importance.

Existing methods support training under diverse resource constraints but often rely on a single shared full-size model for submodel extraction. This shared backbone overlooks client-specific characteristics, leading to structurally similar submodels and performance disparities across clients, as shown in \cref{fig:lim_mask_overlap,fig:lim_client_accuracy}.
Additionally, data heterogeneity poses a critical challenge, as clients with non-IID data may develop divergent representations, resulting in poor generalization and semantic misalignment across the federation.

In this paper, we revisit submodel personalization in FL by addressing two key questions:
\begin{enumerate}[align=left,label=\textbf{Q\arabic*},leftmargin=2.2em,itemsep=1pt, topsep=1pt]
    \item \label{Q1} \textbf{How can we construct personalized submodels that adapt to both the computational constraints and data distribution of each client?}
    \item \label{Q2} \textbf{How can we ensure semantic consistency across clients without relying on any shared public dataset?}
\end{enumerate}

To address \ref{Q1}, we propose client-aware full-size model generation, where each client is assigned a personalized full model conditioned on its compact descriptor. This approach eliminates reliance on a fixed global backbone and reduces redundancy, as many parameters in a shared global model may be irrelevant to specific clients. Building on this, we implement adaptive submodel extraction, where submodels are dynamically selected based on evolving parameter values during training, implicitly capturing parameter importance without additional tracking.

To address \ref{Q2}, we introduce prototype-guided representation alignment, which promotes semantic consistency by aligning class-level prototypes across clients during local training, without requiring any shared public dataset. 

These components form a unified framework that supports personalization in both system-level resource constraints and data heterogeneity.

We call our framework \textit{\textbf{F}ederated \textbf{R}epresentation-\textbf{A}ligned \textbf{M}ulti-scale \textbf{P}ersonalization (FRAMP)}. FRAMP preserves the simplicity of centralized model management by encoding personalization into a single global generator. Our main contributions are summarized as follows:
\begin{itemize}[leftmargin=0.85em,itemsep=2pt, topsep=2pt]
    \item We propose a client-aware model generation mechanism that leverages compact client descriptors to instantiate personalized full-size models, enabling submodel extraction tailored to both computational and data characteristics.
    \item We introduce an adaptive submodel extraction strategy that dynamically encodes parameter importance within the training process, allowing submodels to evolve without additional overhead.
    \item We incorporate a prototype-guided representation alignment strategy that ensures semantic consistency across clients, without requiring any shared public datasets.
    \item Extensive experiments demonstrate that FRAMP achieves robust personalization and generalization, maintaining strong performance even under extreme heterogeneity and unseen client scenarios.
\end{itemize}

%##############################################################
\section{Related Work}
This section highlights related research directions, with a more extensive review in \Cref{app:relatedwork}.

\subsection{Model Heterogeneity in FL}
FL systems often encounter computational heterogeneity, where clients differ significantly in hardware capabilities and resource budgets. A common strategy is to assign more local updates to high-resource clients while limiting those for resource-constrained ones 
\cite{li2020federated,mitra2021linear,shin2022fedbalancer,wu2023deterioration}.
%\cite{li2020federated,wang2020tackling,mitra2021linear,shin2022fedbalancer,wu2023deterioration}.
However, such approaches generally assume that all clients are capable of training the full-size model, which is impractical for severely constrained devices.
To address this, several works explore model customization \cite{lin2020ensemble,zhang2021parameterized,cho2022heterogeneous,zhang2023distill,wu2024fiarse,liu2025model}, enabling clients to train local models adapted to computational capacity. Aggregation across heterogeneous models is often performed via knowledge distillation, which typically requires a shared public dataset \cite{alballa2023first}. However, such requirements limit applicability in decentralized environments where public data is unavailable.

Another approach is model sparsification, where clients prune less important parameters to meet local resource constraints \cite{chen2023efficient,liao2023adaptive,zhou2023every,chen2025advances,zhang2025htfllib}. While effective, these methods may introduce significant computational and memory overhead, limiting their practicality in low-resource scenarios.
In contrast, our framework supports resource-adaptive submodel training without requiring public data and encodes parameter utility implicitly through their values. Furthermore, by coupling model sparsity with personalization, our approach provides a unified solution to both system and data-level heterogeneity, which is rarely addressed simultaneously in existing works.

\subsection{Model Personalization in FL}
To address data heterogeneity in FL, a wide range of \gls{PFL} approaches have been developed. These include local fine-tuning \cite{schneider2021personalization}, regularization-based objectives \cite{hanzely2020lower, yan2024personalized}, model mixing \cite{ma2022layer}, meta-learning \cite{jiang2019improving, lee2024fedl2p}, parameter decompositions \cite{arivazhagan2019federated}.
Other works explore server-side personalization by clustering clients and maintaining multiple global models for different groups \cite{ghosh2020efficient, huang2021personalized}, or by fully decoupling the training of individual models with periodic collaboration \cite{zhang2021parameterized, ye2023personalized, scott2024pefll,liang2025personalized}.
However, most PFL approaches assume structurally identical models across clients, overlooking constraints imposed by device heterogeneity. In contrast, our framework jointly enables resource-aware model adaptation and personalization through a unified architecture.

%####################################
\section{Preliminaries}
\subsection{Problem Formulation}

We consider an FL scenario involving $N$ clients and a central server. Each client $n \in [N]$ has access to a local dataset $\calD_n = \left\{(x_i^n, y_i^n)\right\}_{i=1}^{K_n}$, which may follow a unique, client-specific distribution. The computational capacity of client $n$ is defined by a sparsity budget $\gamma_n \in [0,1]$, indicating the maximum fraction of a full-size model $f_n \in \mathbb{R}^d$ that the client can utilize. This budget $\gamma_n$ varies across clients to reflect their differing resource constraints.

Each client $n$ selects a binary mask $\calM_n\in\left\{0,1\right\}^d$ to extract its submodel, subject to the constraint $\left\|\calM_n\right\|_1\le\gamma_nd$.
Let $\calM=\bigcup_{n\in[N]}\calM_n\in\left\{0,1\right\}^{N\times d}$ denote the collection of all clients' masks. To jointly optimize the model parameters and the masks, the overall objective is defined as:
\begin{align}
\argmin_{\left\{f_n\right\},\calM}
\frac{1}{N}\sum_{n=1}^{N} \E_{(x,y)\sim\calD_n}[\calL_n ((f_n\odot\calM_n)(x),y)], \label{eq:objective_ori}
\end{align}
where $\calL_n$ represents the task-specific loss on client $n$'s local dataset $\calD_n$, and $\odot$ denotes elementwise product. For simplicity, we assume uniform client weighting, though the formulation can naturally extend to non-uniform weights.

\vspace{-0.8mm}
\subsection{Limitations of Existing Approaches}\label{sec:existlimitation}
\vspace{-0.5mm}

Many existing methods rely on a shared full-size global model, where all clients use the same parameters, i.e., $f_1=f_2= \cdots =f_N$, from which they extract local submodels. Each client applies a binary mask $\calM_n$ to select a subset of parameters for local training. These approaches can be categorized into two types: static masking, where the mask remains fixed throughout the training process, and dynamic masking, where the mask is updated across FL rounds.

\begin{figure}[!tb]
\centering
\includegraphics[width=\linewidth]{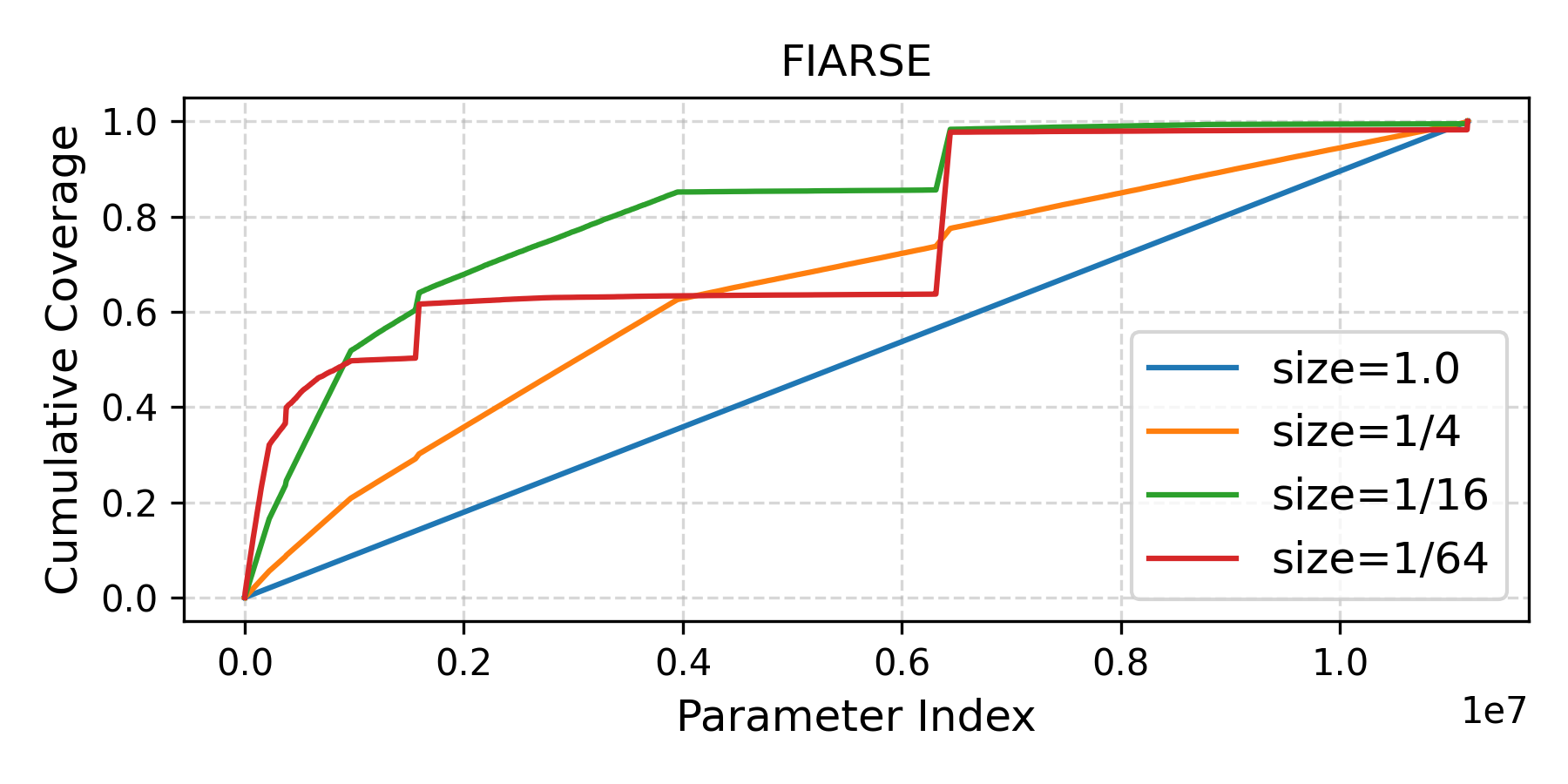}
\vspace{-6.7mm}
\caption{Cumulative distribution of preserved parameters across submodels of different sizes. Each curve represents the cumulative proportion of parameters selected (mask=1).}
\label{fig:lim_mask_overlap}
\vspace{-3.5mm}
\end{figure}

Static methods assign fixed masks to clients, while dynamic approaches allow masks to adapt during training. Some recent methods incorporate importance scores to guide mask selection \cite{liao2023adaptive,chen2023efficient}, alternating between optimizing these scores and updating model parameters. However, such designs often introduce significant computational and storage overhead. Separating parameter updates and mask optimization into distinct steps can also be inefficient.
FIARSE \cite{wu2024fiarse} mitigates this by ranking the parameters based on their magnitude and pruning accordingly, avoiding explicit mask optimization. However, it still relies on a globally shared full model and applies a fixed global ranking for submodel extraction across all clients. This approach has two major drawbacks:
\begin{itemize}[leftmargin=0.85em,itemsep=1pt, topsep=1pt]
    \item \textbf{Limited Structural Diversity:} Since all clients prune from the same global full model using a universal importance ranking, the extracted submodels are structurally similar, regardless of variations in client data distributions.
    \item \textbf{Client-Agnostic Importance Estimation:} Parameter importance is determined without accounting for client-specific characteristics, often yielding submodels that fail to adapt to client data.
\end{itemize}
As shown in \cref{fig:lim_mask_overlap}, the cumulative coverage plots across four sparsity levels $\gamma_n$ reveal that FIARSE concentrates mask activations on a similar subset of model parameters across clients within each model-size group.
For example, the 1/64 submodels (red line) place more than 60\% of mask activations within the first 20\% of the parameter index range, while many later parameters have low selection probability. This behavior stems from all clients selecting submodels based on a shared importance score ranking. Although FIARSE supports dynamic submodel sampling in each round, parameters with low importance scores are seldom chosen, resulting in limited training opportunities for these parameters. Consequently, a significant portion of the model remains underutilized, leading to high structural similarity and reduced diversity among submodels. Additional discussion is in \Cref{app:furthercomparison}.

%#############################
\begin{figure}[!tb]
\centering
\includegraphics[width=\linewidth]{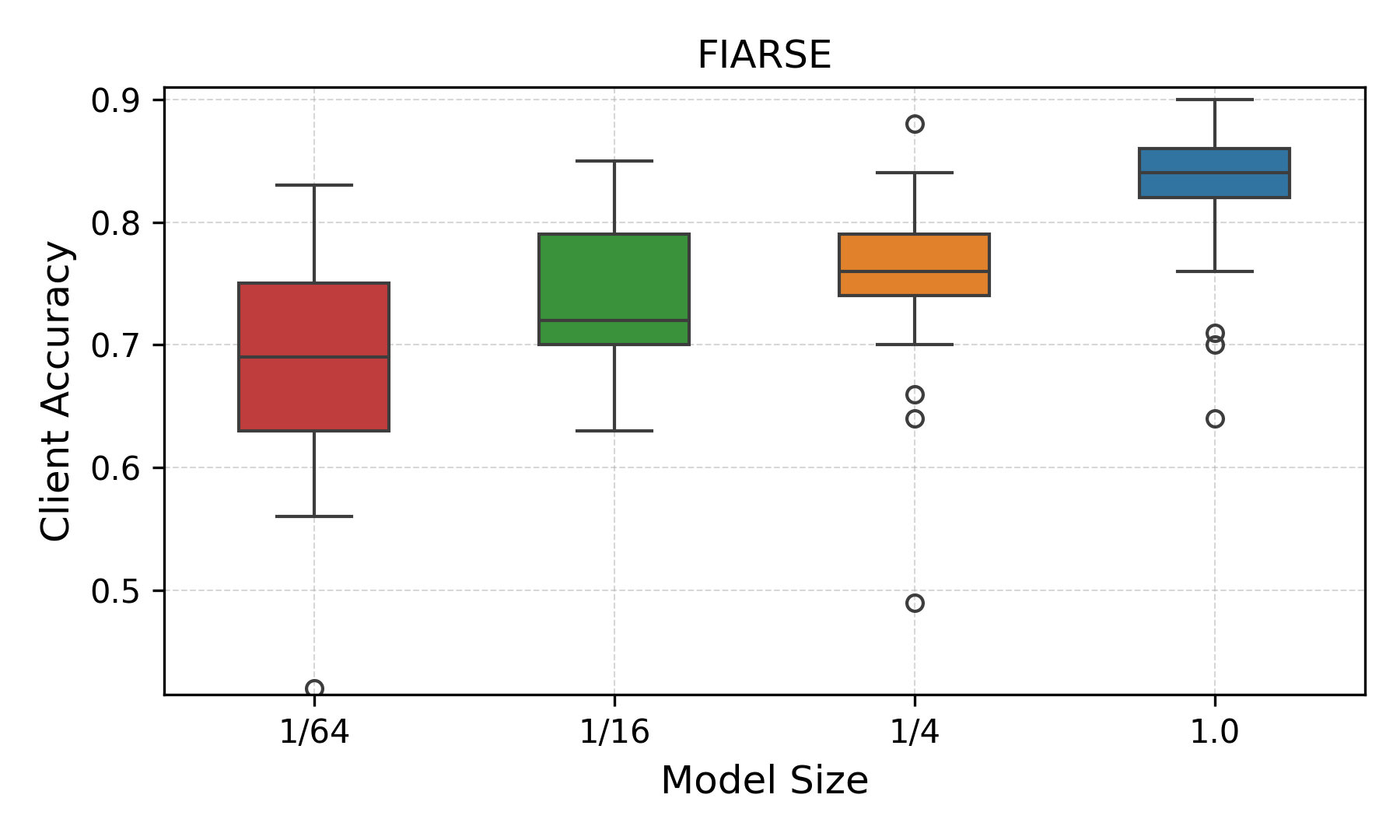}
\vspace{-6.7mm}
\caption{Client accuracy distribution of submodels across different sizes. Each box represents the variation in accuracy among clients.}
\label{fig:lim_client_accuracy}
\vspace{-3.5mm}
\end{figure}
%%%%%%%%%%%%%%%%%%%%%%%%%%%%%

Furthermore, \cref{fig:lim_client_accuracy} highlights significant performance disparities among clients, particularly those with smaller model sizes, suggesting that the extracted submodels fail to address client-specific needs. Additionally, most existing methods overlook the challenge of representation misalignment, where clients learn inconsistent class prototypes due to model heterogeneity and non-IID data \cite{liu2025model}, ultimately compromising generalization performance.

%%%%%%%%%%%%%%%%%%%%%%%%%%%
\begin{figure*}[!t] 
\centering
\includegraphics[width=\textwidth]{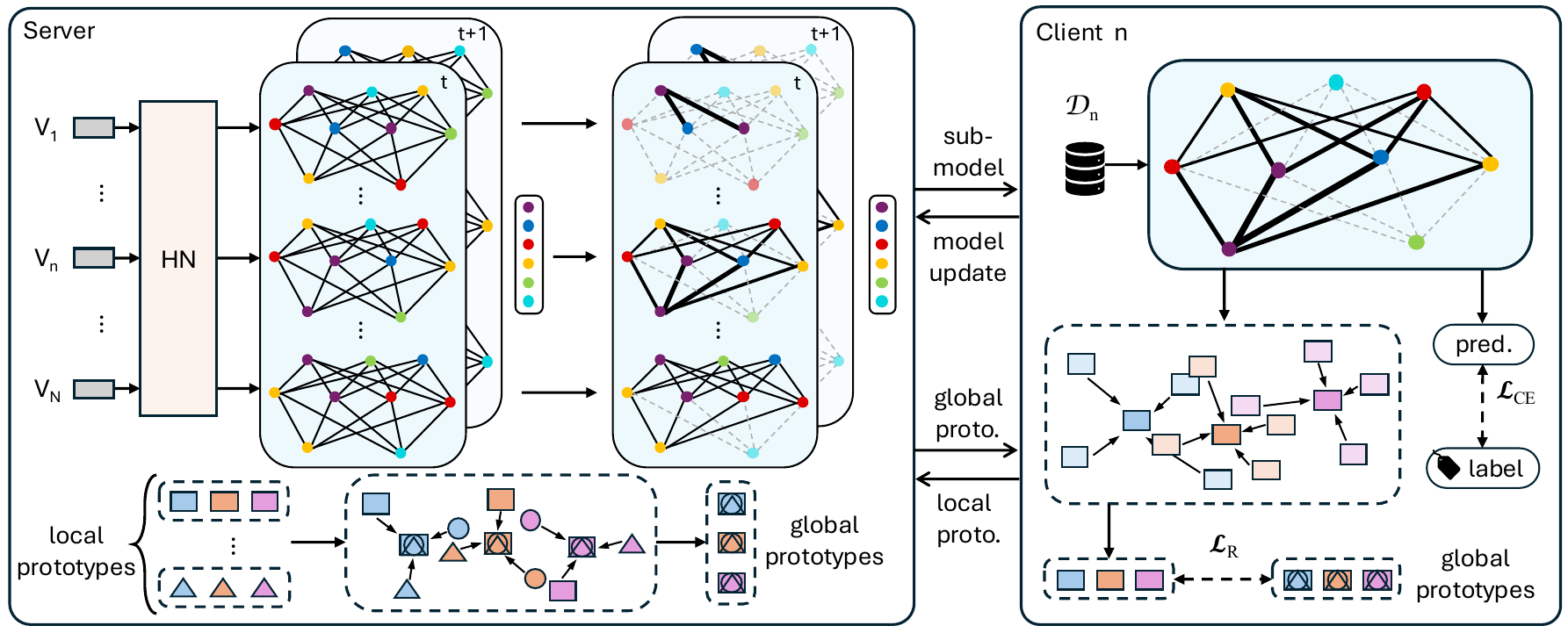}
\caption{Overview of the FRAMP framework. The server generates personalized full models using HN and extracts submodels via dynamic masking for each client. Clients train these submodels on local data and align class-level prototypes with server-aggregated global prototypes to ensure semantic consistency. Updates are used to refine the HN and global prototypes.}
\label{fig:wholeframework}
\vspace{-1.5mm}
\end{figure*}
%%%%%%%%%%%%%%%%%%%%%%%%%%%

%%%%%%%%%%%%%%%%%%%%%%%%%%%%%%%%%%%%%%%%%%%%%%%%%%%%%%%%%%%%%%%%%%%%%%
\vspace{-2mm}
\section{Methodology}
\vspace{-1mm}

FRAMP proceeds in three key stages, which are repeated during every communication round.
An overview of the framework is shown in \cref{fig:wholeframework}.
\begin{itemize}[leftmargin=0.85em,itemsep=1pt, topsep=1pt]
    \item \textbf{Stage 1 – Client-Aware Full-Size Model Generation} (Section~\ref{sec:fullmgenerate}): The server generates a personalized full-size model for each client $n$ by feeding its client-specific vector $\bv_{n}$ into the server model $H(\cdot;\bm{\varphi})$, which outputs model parameters tailored to the client.
    \item \textbf{Stage 2 – Adaptive Submodel Extraction} (Section~\ref{sec:submextract}):  
    Each client adaptively derives a sparse submodel from the generated full-size model based on importance scores, constrained by its local computation budget $\gamma_n$.
    \item \textbf{Stage 3 – Representation Alignment} (Section~\ref{sec:repalign}):  
    Clients train their submodels on local data while aligning class-level representations with global prototypes to maintain semantic consistency across heterogeneous clients.
\end{itemize}

At initialization, each client $n$ uploads its vector $\bv_{n}$ to the server, serving as the input for model generation. Stages 1 to 3 are then repeated until the final communication round $R$. In summary, FRAMP leverages client-specific information to generate personalized full-size models, dynamically extracts sparse submodels based on parameter importance, and promotes semantic alignment during local training. The full procedure is detailed in \cref{alg:framp} in \Cref{app:algorithm}.

%%%%%%%%%%%%%%%%%%%%%%%%%%%%%%%%%%%%%%%%%%%%%%%%%%%%%%%%%%%%%%%%%%%%%%
\vspace{-0.5mm}
\subsection{Client-Aware Full-Size Model Generation}\label{sec:fullmgenerate}
\vspace{-0.5mm}

To address the limitations of global model sharing, we generate a personalized full-size model $f_n(\cdot;\bomega_n)$ for each client $n$, where the parameters $\bomega_n$ are produced by an \gls{HN} $H(\cdot;\bm{\varphi})$ conditioned on a client-specific descriptor $\bv_{n}\in \mathbb{R}^l$. This design enables per-client model generation in a scalable and unified framework, while remaining consistent with the general problem formulation.

At initialization, each client computes its descriptor $\bv_{n}$ by aggregating embeddings of local data instances. Specifically, we employ a feature extractor $\phi$ to encode individual samples and take the average over the local dataset, i.e.,  $\bv_{n}=\frac{1}{\left|K_n\right|}{\textstyle \sum_{(x,y)\in \calD_n}}\phi(x)$. 
The descriptor can be fixed or updated during training, provided that a meaningful client representation is available a priori.

The HN maps each client descriptor $\bv_{n}$ to the corresponding model parameters:
\begin{align}\label{eq:theta_n}
\bomega_n := H(\bv_{n};\bm{\varphi}),
\end{align}
defining the full-size model $f_n$ for client $n$. By mapping each descriptor to a personalized full-size model, the HN ensures that all client-specific models reside on a shared manifold in the parameter space \cite{shamsian2021personalized}. This approach allows each full-size model to adapt to the unique characteristics of its client's data while maintaining a structured parameter space that promotes generalization. Sharing the HN parameters $\bm{\varphi}$ enables efficient knowledge transfer across clients, balancing personalization with collaborative learning.

%%%%%%%%%%%%%%%%%%%%%%%%%%%%%%%%%%%%%%%%%%%%%%%%%%%%%%%%%%%%%%%%%%%%%%
\vspace{-0.5mm}
\subsection{Adaptive Submodel Extraction}\label{sec:submextract}
\vspace{-0.5mm}

Prior studies \cite{mostafa2019parameter,jayakumar2020top,wu2024fiarse} have demonstrated that the magnitude of model parameters can serve as a useful proxy for their importance. By controlling the number of active parameters in a submodel, we constrain its computation and storage cost to meet client-specific budgets. To this end, we define a client-specific threshold such that only parameters with absolute values exceeding this threshold are retained in the submodel. Formally, the binary mask in Eq.~(\ref{eq:objective_ori}) becomes a deterministic function of the model parameters:
\begin{align}
&\argmin_{\left\{f_n\right\}}
\frac{1}{N}\sum_{n=1}^{N} \E_{(x,y)\sim\calD_n}[\calL_n ((f_n\odot\calM_n(\bomega_n))(x),y)], \nonumber \\
&\mathrm{where}\quad \calM_n(\bomega_n)=
\begin{cases}
1, & \text{if}\ \left|\bomega_n\right|\ge \btheta_n, \\
0, & \text{if}\ \left|\bomega_n\right|<  \btheta_n.
\end{cases}
\label{eq:objective_subm}
\end{align}
Here, $\calM_n(\cdot)$ denotes the mask function for client $n$, determined by a threshold $\btheta_n$ such that $\left\|\calM_n(\bomega_n)\right\|_1\le\gamma_nd$. 
The threshold $\btheta_n$ is determined according to each client's local computation budget. Specifically, we apply a $\mathrm{TopK}_{\gamma_n}(\cdot)$ operation to retain the top $\gamma_nd$ entries in $\left|\bomega_n\right|$, where $\btheta_n$ is set as the minimum value among the selected entries.

This formulation reduces submodel extraction to a thresholding operation over parameter magnitudes. 
As a result, optimizing model parameters implicitly determines mask and forms submodel, effectively prioritizing parameters with higher importance without requiring explicit mask updates.

%%%%%%%%%%%%%%%%%%%%%%%%%%%%%%%%%%%%%%%%%%%%%%%%%%%%%%%%%%%%%%%%%%%%%%
\subsection{Representation Alignment}\label{sec:repalign}

To enhance semantic consistency across clients, we align class-level representations derived from the local model. For this purpose, we decompose each client model $f_n$ into two components: an \textit{encoder} $e_n:\mathbb{R}^k\to \mathbb{R}^h$, which maps inputs to $h$-dimensional latent representations, and a \textit{prediction head} $g_n:\mathbb{R}^h\to \mathbb{R}^C$, which produces logits for $C$ classes.
Each client $n$ computes a local prototype $\bP_n^c\in \mathbb{R}^h$ for class $c$ by averaging the encoder outputs over the local samples belonging to that class as
$\bP_n^c=\frac{1}{\left|\calX_n^c\right|}\sum_{x\in \calX_n^c}e_n(x)$, where $\calX_n^c$ denotes the set of client $n$'s local samples with label $c$.

After local training, each participating client uploads its set of local prototypes $\left\{\bP_n^c\right\}_{c=1}^C$ to the server. The server aggregates these to obtain a global prototype for each class via $\bP^c=\frac{1}{N}\sum_{n=1}^{N}\bP_n^c$, and broadcasts the global prototype set $\bP_g=\left\{\bP^c\right\}_{c=1}^C$ back to the clients.

To leverage global semantic knowledge during local updates, a prototype alignment regularization term is added to the local loss function. This term encourages the client's local representations to stay close to the global counterparts as
\begin{align}
\calL_n=\calL_{CE}+\lambda \calL_R,\ \calL_R=\sum_{c=1}^{C} \mathrm{dist}(\bP_n^c,\bP^c),
\end{align}
where $\calL_{CE}$ is the standard cross-entropy loss, $\calL_R$ is the prototype alignment loss, $\lambda$ is a balancing coefficient, and $\mathrm{dist}(\cdot,\cdot)$ is a distance function (e.g., Euclidean distance).

%%%%%%%%%%%%%%%%%%%%%%%%%%%%%%%%%%%%%%%%%%%%%%%%%%%%%%%%%%%%%%%%%%%%%%
\subsection{Learning Procedure}

With the personalized model generation and submodel extraction mechanisms, the overall learning objective is reformulated as:
% \begin{align}
% \argmin_{\bm{\varphi}}
% \frac{1}{N}\sum_{n=1}^{N} \E_{(x,y)\sim\calD_n}[\calL_n ((H(\bv_{n};\bm{\varphi})\odot\calM_n(\bomega_n))(x),y)].
% \label{eq:objective_hn}
% \end{align}
\begin{align}
\resizebox{1.05\linewidth}{!}{$
\displaystyle
\argmin_{\bm{\varphi}}
\frac{1}{N}\sum_{n=1}^{N} \E_{(x,y)\sim\calD_n}[\calL_n ((H(\bv_{n};\bm{\varphi})\odot\calM_n(\bomega_n))(x),y)].
\label{eq:objective_hn}$}
\end{align}
After receiving the submodel $f_n\odot\calM_n(\bomega_n)$ from the server in the current communication round, client $n$ performs $T$ steps of local training on its private dataset $\calD_n$, updating only the active submodel parameters as $\hat\bomega_n^{t+1} = \hat\bomega_n^t - \alpha\nabla_{\hat\bomega_n}\calL_n(\hat\bomega_n^t)$, where $\hat\bomega_n$ denotes the submodel parameters selected by the mask.
After $T$ steps, the client transmits the update $\Delta \hat\bomega_n = \hat\bomega_n^T - \hat\bomega_n^0$ to the server. 
% \blue{which corresponds to the projected update on the active parameter subset determined by the mask}. 
The server then updates the HNs parameters $\bm{\varphi}$ as $\bm{\varphi}:=\bm{\varphi}-\beta\Delta \bm{\varphi}$, where $\Delta\bm{\varphi}=(\nabla_{\bm{\varphi}}\hat\bomega_n)^T\Delta\hat\bomega_n, \nabla_{\bm{\varphi}}\hat\bomega_n=\calM_n\odot\nabla_{\bm{\varphi}}\bomega_n$.
This formulation allows FRAMP to jointly optimize the server-side model and client-specific submodels through end-to-end gradient-based updates.

%##########################################################################
\begin{table*}[!ht]
\caption{Test accuracy of submodels across four sizes on CIFAR-10, CIFAR-100, and ogbn-arxiv.}
\label{tb:localacc}
\vspace{-1mm}
\centering
\resizebox{\textwidth}{!}{
\begin{tabular}{lcccccccccccccc}
\toprule
\multirow{2}{*}{Method} & \multicolumn{6}{c}{CIFAR-10} & \multicolumn{6}{c}{CIFAR-100} & \multicolumn{2}{c}{ogbn-arxiv}\\
\cmidrule(lr){2-7} \cmidrule(lr){8-13} \cmidrule(lr){14-15}
&Local & 1/64 & 1/16 & 1/4 & 1.0 &Union & Local & 1/64 & 1/16 & 1/4 & 1.0 &Union &Local &Union\\
\midrule
HeteroFL & 68.88 & 60.24 & 69.32 & 72.18 & 73.76 & 66.05  & 31.75 & 27.24 & 29.80 & 33.52 & 36.44 & 30.67 &31.53  &31.26\\
FedRolex & 67.18 & 54.60 & 64.96 & 70.08 & 79.08 & 65.98  & 31.67 & 21.00 & 30.84 & 36.44 & 38.40 & 29.89 &28.26  &27.98 \\
ScaleFL  & 72.10 & 69.04 & 71.64 & 70.08 & 77.64 & 67.37  & 39.69 & 36.16 & 40.48 & 42.56 & 39.56 & 37.56 &41.53 &37.77 \\
FIARSE   & 77.04 & 73.12 & 77.20 & 77.24 & 82.04 & 73.75  & 41.76 & 39.12 & 43.24 & 43.72 & 40.96 & 38.63 &46.18 &41.53 \\
FRAMP    &\textbf{79.11} &\textbf{78.20} &\textbf{78.56} &\textbf{77.40} &\textbf{82.28} &\textbf{75.65} &\textbf{42.95} &\textbf{43.00} &\textbf{44.08}  &\textbf{43.72} &\textbf{41.00} &\textbf{40.26} &\textbf{48.00} &\textbf{44.34}\\
\bottomrule
\end{tabular}
}
\vspace{-2mm}
\end{table*}
\begin{figure*}[!t] 
\centering
\includegraphics[width=\textwidth]{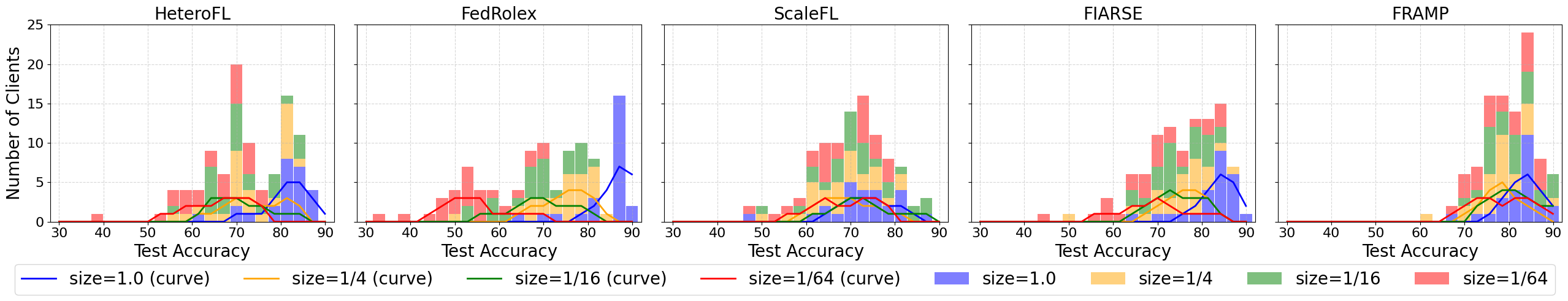} 
\vspace{-5mm}
\caption{Histograms on CIFAR-10 showing client counts at different test accuracy levels under four submodel sizes for each baseline. The curves depict the accuracy distribution for each model size.}
\label{fig:histogram}
\vspace{-2mm}
\end{figure*}
\begin{figure*}[!t]
\centering
\begin{subfigure}{0.31\textwidth}
    \centering
    \includegraphics[width=\columnwidth]{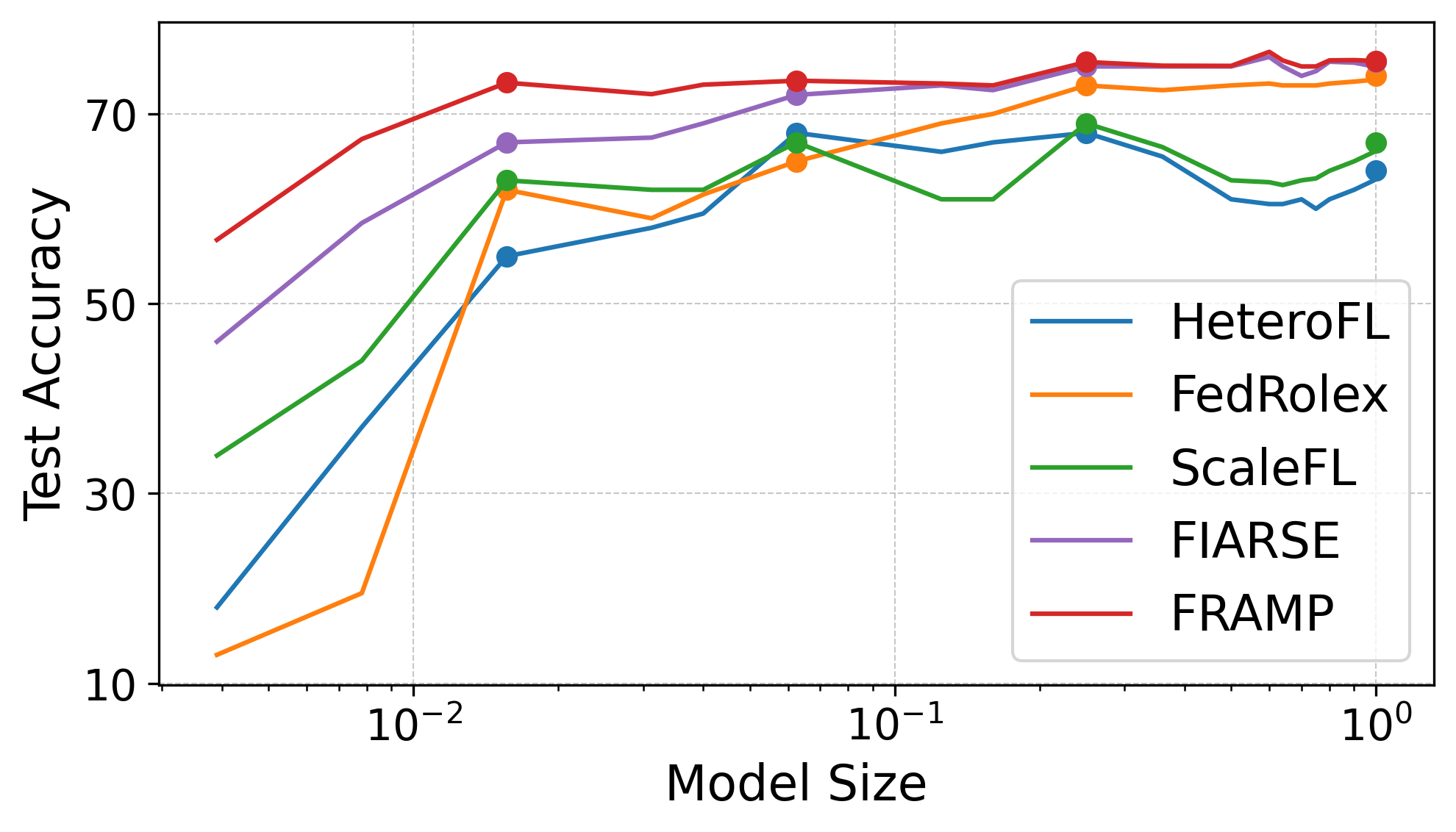}
    \vspace{-4mm}
    \caption{Submodel accuracy.}
    \label{fig:diffsubmodel}
\end{subfigure}
\hfill
\begin{subfigure}{0.31\textwidth}
    \centering
    \includegraphics[width=\columnwidth]{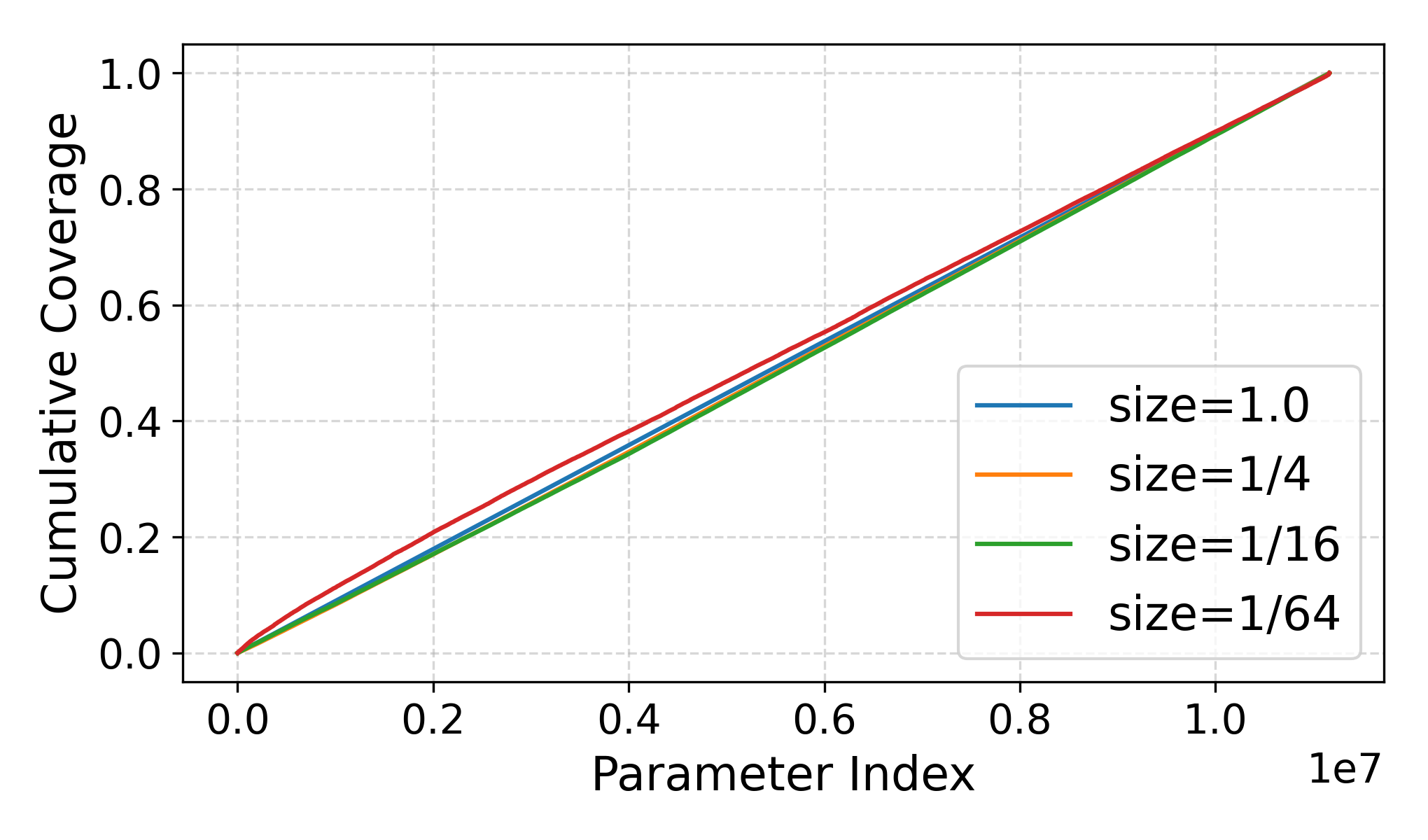}
    \vspace{-4mm}
    \caption{Cumulative coverage of masks.}
    \label{fig:ab_mask}
\end{subfigure}
\hfill
\begin{subfigure}{0.35\textwidth}
    \centering
    \includegraphics[width=\columnwidth]{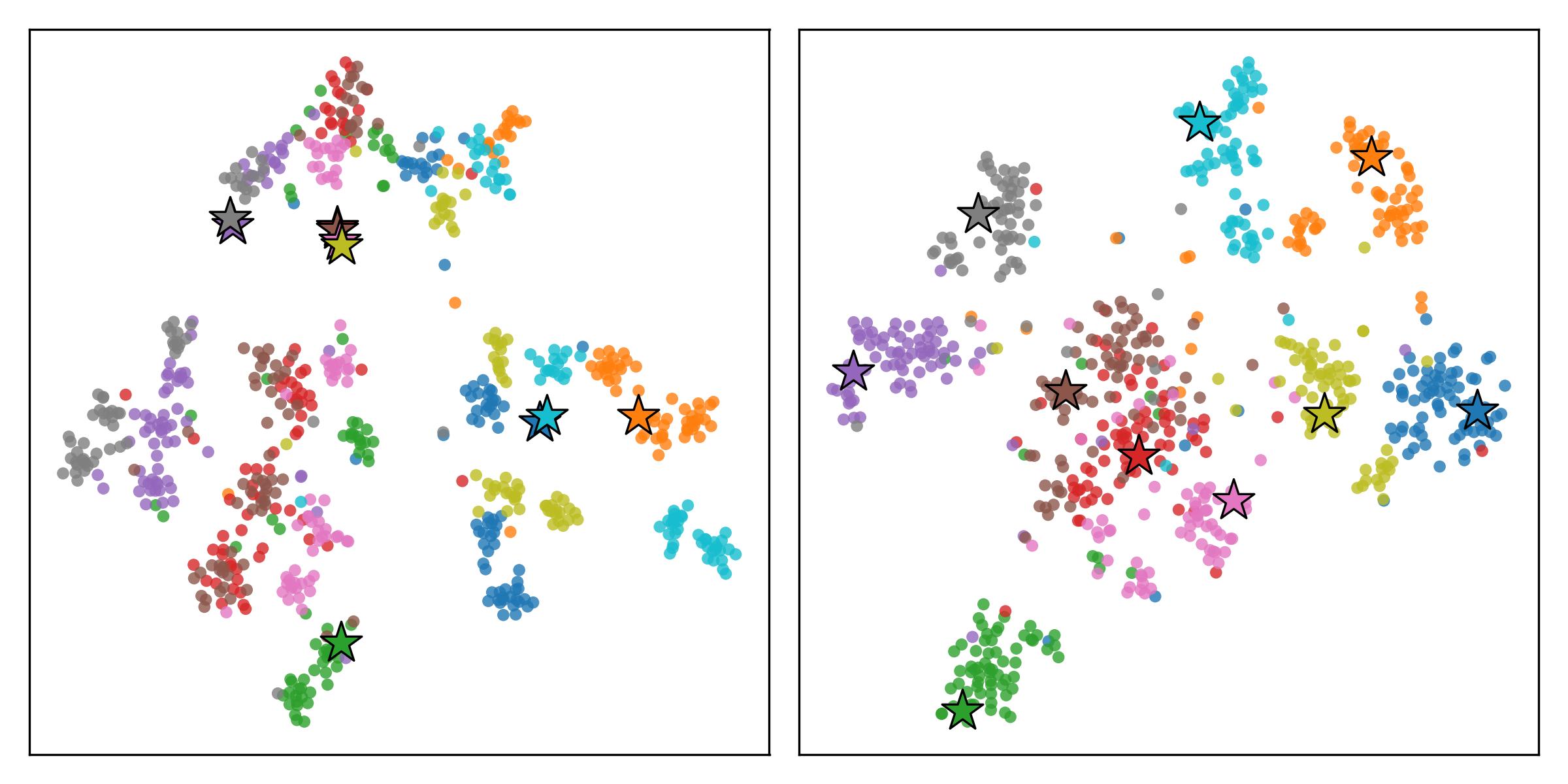}
    \vspace{-4mm}
    \caption{t-SNE visualization.}
    \label{fig:ab_tsne}
\end{subfigure}
\vspace{-0.5mm}
\caption{(a) Test accuracy of submodels with different sizes for all baselines, evaluated on the union test set of CIFAR-10. (b) Cumulative coverage of parameter masks for different submodel sizes in FRAMP, showing a more uniform and efficient utilization of full-model parameters across submodels, compared to \cref{fig:lim_mask_overlap}. (c) t-SNE visualization of local prototypes (dots, one per class per client) and global prototypes (stars) in FRAMP on CIFAR-10. Each color represents a class.}
\vspace{-3mm}
\end{figure*}
%%%%%%%%%%%%%%%%%%%%%%%%%%%%%%%%%%%%%%%%%%%%%%%%%%%%%%%%%%%%%%%

%%%%%%%%%%%%%%%%%%%%%%%%%%%%%%%%%%%%%%%%%%%%%%%%%%%%%%%%%%%%%%%%%%%%%%
\section{Experiments}
\subsection{Experimental Settings}

\textbf{Datasets}\hspace{0.7em}
We evaluate FRAMP on image classification (CIFAR-10, CIFAR-100 \cite{krizhevsky2009learning}) and node classification on graph-structured data (ogbn-arxiv \cite{hu2020open}). Following \cite{wu2024fiarse}, CIFAR-10 and CIFAR-100 are partitioned into 100 clients using a Dirichlet distribution with concentration parameter $\alpha = 0.3$, while ogbn-arxiv is partitioned into disjoint subgraphs per client as in \cite{baek2023personalized}. Further partitioning details are provided in \Cref{app:dataset}.

\textbf{Baselines}\hspace{0.7em}
We compare FRAMP with state-of-the-art FL methods supporting model heterogeneity, including static submodel extraction approaches HeteroFL \cite{diao2020heterofl}, ScaleFL \cite{ilhan2023scalefl}, and dynamic strategies FedRolex \cite{alam2022fedrolex}, FIARSE \cite{wu2024fiarse}. All baselines extract submodels from a shared global backbone, without client-specific personalization.

\textbf{System Heterogeneity}\hspace{0.7em}
To simulate system-level constraints, we vary the sparsity ratio $\gamma_n$, which specifies the proportion of the full model that client $n$ can store and train.
We evaluate four capacity levels, $\gamma'=\left\{1/64,1/16,1/4,1.0\right\}$, with clients evenly assigned to each level. 
Our framework can be easily extended to other client distributions or finer-grained capacity groups.

\textbf{Implementation Details}\hspace{0.7em}
The client participation ratio is set to 10\% by default. Training is run for 800 communication rounds on image tasks and 200 rounds on graph tasks. 
We use ResNet-18 for image tasks and a four-layer GCN for graph tasks. The HNs in FRAMP is implemented as a two-layer MLP.
Reported results are averaged over three random seeds. All baseline results are obtained on the best hyperparameter settings as in \Cref{app:hyperparameters}.

%%%%%%%%%%%%%%%%%%%%%%%%%%%%%%%%%%%%%%%%%%%%%%%%%%%%%%%%%%%%%%%%%%%%%%
\subsection{Performance Comparison}

We evaluate submodel performance on each client's local test dataset, as summarized in \cref{tb:localacc}. Columns “1/64” to “1.0” report the average accuracy grouped by model size $\gamma_n$, while the “Local” column shows the overall average across all clients. 
FRAMP consistently outperforms all baselines, with particularly notable gains for resource-limited clients. 
In the smallest submodel setting (1/64), FRAMP achieves substantial improvements across all datasets, suggesting that combining dynamic masking with HN-based full-model personalization effectively guides resource-constrained models toward the most informative parameters.

\cref{fig:histogram} illustrates the distribution of client accuracies across different submodel sizes.
FedRolex, with its rolling-based strategy, exhibits a dispersed and inconsistent distribution, highlighting challenges in optimizing local submodel performance. HeteroFL shows notable disparities across model sizes, with smaller submodels performing substantially worse than full-size models. FIARSE achieves strong results for larger models but suffers from greater variability and a pronounced long-tail effect for smaller models, indicating reduced accuracy for resource-constrained clients. 

In contrast, FRAMP exhibits a compact, right-shifted distribution across all submodel sizes, reflecting consistently high accuracy regardless of model capacity. The smoothed curves for different sizes largely overlap, indicating strong inter-size consistency. Overall, FRAMP delivers balanced, stable, and robust performance across all submodel sizes, even under extreme resource constraints. 

These advantages stem from FRAMP’s balanced parameter utilization and representation alignment. \cref{fig:ab_mask} shows more even mask activations across parameters. \cref{fig:ab_tsne} demonstrates representation alignment, with local prototypes (dots) relatively dispersed at an earlier stage (left) converging toward global prototypes (stars) at a later stage (right), forming tighter class clusters.

%%%%%%%%%%%%%%%%%%%%%%%%%%%%%%%%%%%%%%%%%%%%%%%%%%%%%%%%%
\begin{figure*}[!ht] 
\centering
\includegraphics[width=\textwidth]{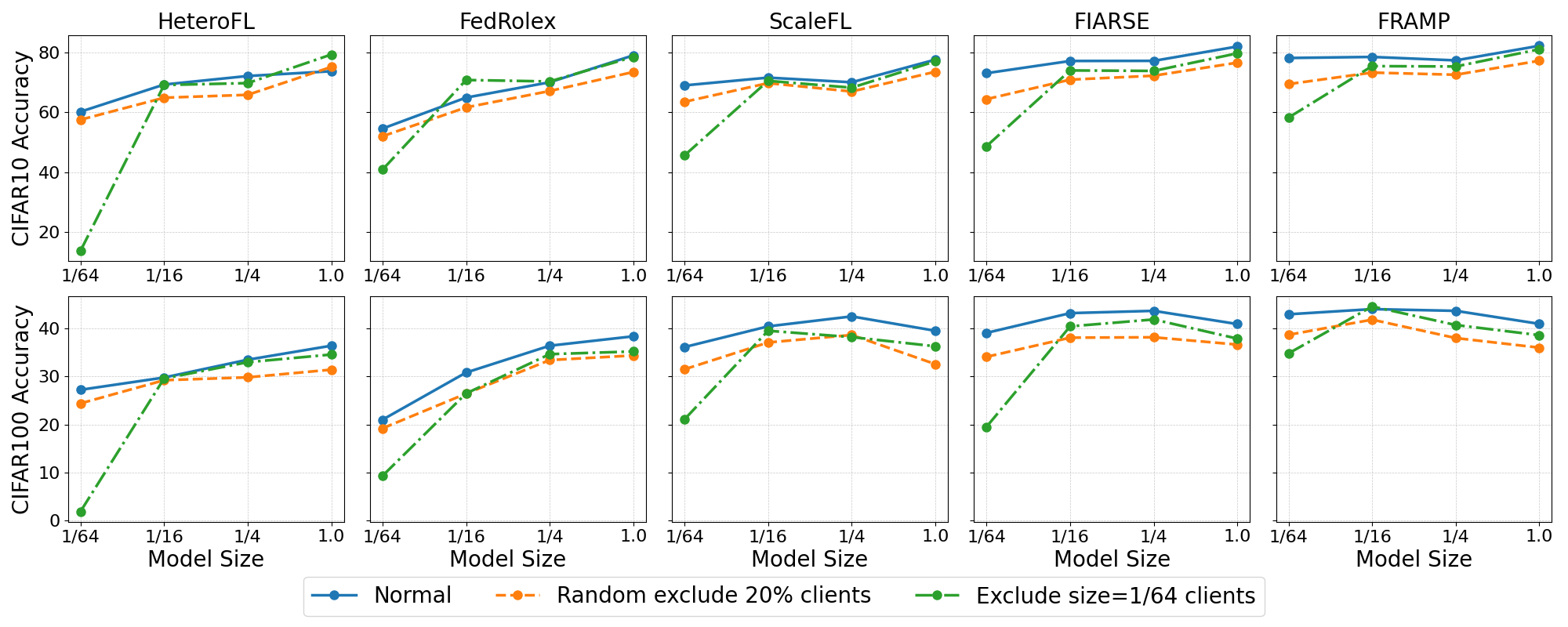} 
\vspace{-5mm}
\caption{Test accuracy of submodels (across four model sizes) on CIFAR-10 and CIFAR-100, evaluated under three training settings: (1) training with all clients; (2) randomly excluding 20\% of clients from each model size group; (3) excluding only clients with the smallest model size ($\gamma=1/64$).}
\label{fig:newclients}
\vspace{-3mm}
\end{figure*}
%%%%%%%%%%%%%%%%%%%%%%%%%%%%%%%%%%%%%%%%%%%%%%%%%%%%%%%%%
%%%%%%%%%%%%%%%%%%%%%%%%%%%%%%%%%%%%%%%%%%%%%%%%%%%%%%%%%%%%

\vspace{-1mm}
\subsection{Generalization on Union Test Set}
\vspace{-0.5mm}

Evaluating on the union test set provides a measure of global generalization beyond individual client distributions. We test submodels of different sizes on a union set formed by merging all clients’ local test data. For baselines that train a global full-size model, submodels are extracted directly from the trained model. Since FRAMP does not explicitly train such a model, we synthesize one via the trained HNs using the average of all client descriptors, and extract submodels in the same manner.

The "Union" column in \cref{tb:localacc} reports the average performance of submodels (across sizes that participated in training) evaluated on the union test set. FRAMP consistently outperforms all baselines.

We further consider a more challenging scenario where unseen clients require submodels of sizes not encountered during training. To simulate this, we extract both larger and smaller submodels and evaluate them on the union test set. As illustrated in \cref{fig:diffsubmodel}, FRAMP consistently outperforms all baselines. While larger submodels generally lead to better performance for all methods, the baselines, particularly static ones such as HeteroFL and ScaleFL, suffer significant degradation under size shifts. In contrast, FRAMP maintains strong adaptability even under extreme deviations.

%%%%%%%%%%%%%%%%%%%%%%%%%%%%%%%%%%%%%%%%%%%%%%%%%%%%%%%%%%%%%%
% \vspace{-2mm}
\subsection{Generalization to New Clients}
% \vspace{-1mm}

While the previous section evaluates submodels on a union test set using clients who participated in training, this section focuses on generalization to entirely unseen clients. 
In real-world deployments, new clients with distinct data may arrive after training, necessitating personalized models despite not being involved in the optimization process.

To simulate this scenario, we randomly exclude 20\% of clients from each model size group during training and evaluate the resulting models directly on their local test sets. For baselines, submodels are extracted from the trained global model and evaluated on the new clients. For FRAMP, we freeze the model generator $H(\cdot;\bm{\varphi})$ and initialize personalized models using the descriptors of the new clients. \cref{fig:newclients} shows FRAMP maintains comparable accuracy on unseen clients with minimal degradation.

We further evaluate an extreme case where all clients with model size $\gamma_n=1/64$ are excluded from training. This setting examines generalization to capacity constraints entirely absent during optimization. As shown in \cref{fig:newclients}, FIARSE exhibits a sharp performance drop on excluded $1/64$ clients, and this exclusion also adversely impacts performance on other model sizes. Although FRAMP also experiences a decline, it consistently outperforms other baselines, demonstrating strong robustness to missing capacity profiles.

%%%%%%%%%%%%%%%%%%%%%%%%%%%%%%%%%%%%%%%%%%%%%%%%%%%%%%%%%%%%%
% \vspace{-2mm}
\subsection{Ablation Studies}
\vspace{-0.7mm}
\textbf{Impact of Model Personalization and Representation Alignment}\hspace{0.7em}
We conduct an ablation study by selectively disabling individual modules. \cref{tb:ab_components} reports the performance of four variants. To isolate semantic alignment, we remove prototype loss by setting $\lambda=0$. To assess personalization, we replace \glspl{HN} with a shared full-size model while retaining the same masking strategy. Finally, we remove both. The results show that disabling either module consistently reduces performance. While the average accuracy drop is modest, removing the alignment loss notably increases class-wise accuracy variance. These findings indicate that the two modules provide complementary benefits: model personalization improves adaptability to heterogeneous clients, while prototype alignment fosters semantic consistency under non-IID distributions.

%%%%%%%%%%%%%%%%%%%%%%%%%%%%%%%%%%%%%%
\vspace{-0.7mm}
\textbf{Submodel Extraction Strategy Analysis}\hspace{0.7em}
We compare different strategies for extracting submodels under resource constraints. Specifically, we evaluate three approaches: (i) TopK-based masking by parameter magnitude (adopted in FRAMP), (ii) layerwise TopK masking that selects within each layer, and (iii) a learned projection method that adds an MLP to the full model output from \glspl{HN}.

As shown in \cref{tb:ab_components}, global TopK masking consistently performs best across all model sizes. Layerwise TopK performs worse, since independent selection within each layer ignores global parameter importance and leads to suboptimal allocation. The MLP-based approach, though end-to-end, suffers from unstable training and degraded performance. This may be due to the difficulty of learning structured sparsity and controlling sparsity levels.
In contrast, global TopK provides a simple yet effective way to enforce budget constraints while retaining essential parameters.

%%%%%%%%%%%############################# 
\begin{table}
\caption{Ablation study on CIFAR-10.}
\label{tb:ab_components}
\vspace{-1mm}
\centering
\resizebox{\linewidth}{!}{
\begin{tabular}{lcccccc}
\toprule
Method &Local & 1/64 & 1/16 & 1/4 & 1.0 &Union \\
\midrule
FRAMP                           &79.11 &78.20 &78.56 &77.40 &82.28 &75.65 \\
w/o RA\textsuperscript{(a)}     &74.49 &75.16 &74.12 &72.44 &76.24 &72.40 \\
w/o Per\textsuperscript{(b)}    &75.21 &70.68 &74.40 &75.44 &80.32 &71.17 \\
w/o Both\textsuperscript{(c)}   &73.97 &68.68 &74.40 &74.72 &78.08 &70.84 \\
\midrule
Layerwise\textsuperscript{(d)}  &67.79 &63.00 &69.04 &67.52 &71.60 &65.46 \\
MLP\textsuperscript{(e)}        &67.12 &62.92 &66.96 &65.68 &72.92 &64.12 \\
\midrule
Onehot\textsuperscript{(f)}     &76.52 &76.40 &77.60 &74.04 &78.04 &74.27 \\
Update\textsuperscript{(g)}     &78.69 &77.02 &78.00 &77.40 &82.32 &75.01 \\
\bottomrule
\end{tabular}
}
\vspace{4pt}
\begin{minipage}{0.95\linewidth}
\small
\textsuperscript{(a)} Remove the representation alignment loss $\mathcal{L}_R$. \\
\textsuperscript{(b)} Replace personalized full model with a shared global model. \\
\textsuperscript{(c)} Remove full model generation and representation alignment. \\
\textsuperscript{(d)} Use layer-wise $\mathrm{TopK}$ masking instead of global $\mathrm{TopK}$. \\
\textsuperscript{(e)} Replace masking with an MLP-based submodel generator.\\
\textsuperscript{(f)} Use onehot vectors as descriptors.\\
\textsuperscript{(g)} Update the descriptors during training.
\end{minipage}
\vspace{-5mm}
\end{table}

%%%%%%%%%%%#############################

%%%%%%%%%%%%%%%%%%%%%%%%%%%%%%%%%%%%%%
\textbf{Descriptor Strategy Analysis}\hspace{0.7em}
We compare three approaches: (i) using a randomly initialized feature extractor to generate fixed descriptors (used in FRAMP), (ii) assigning each client a one-hot vector, and (iii) a trainable feature extractor and updating descriptors every 10 communication rounds.  
As shown in \cref{tb:ab_components}, one-hot vectors perform worst, as lacking client-specific information. (i) and (iii) yield similar accuracy, but (iii) incurs higher client computation due to training the feature extractor.

%%%%%%%%%%%%%%%%%%%%%%%%%%%%%%%%%%%%
Sensitivity of $\lambda$, controlling the weight of the alignment loss, is reported in \Cref{app:sensityhyper}. 

%%%%%%%%%%%%%%%%%%%%%%%%%%%%%%%%%%%%%%%%%%%%%%%%%%%%%%%%%%%%%%% 
\subsection{Further Discussions}

\textbf{Computational Overhead}\hspace{0.7em}
The HNs operate entirely on the server, adding no extra computation for clients compared to standard heterogeneous FL baselines. At initialization, each client sends its descriptor to server once, which remains fixed throughout training, incurring no extra communication. The only additional cost during training is prototype exchange, which adds negligible overhead.
Additional discussion is provided in \Cref{app:overhead}.

\cref{tab:runtime} reports the server-side training time per communication round. FRAMP introduces a modest overhead in the first round due to the initialization of the model generator, but maintains comparable training time to existing baselines in subsequent rounds.

%%%%%%%%%%%%%%%%%%%%%%%%%

\textbf{Privacy Discussion}\hspace{0.7em}
FRAMP transmits three types of information during training: submodels, masked gradients, and class-level prototypes. The former two are standard in FL pipelines, while prototype sharing aligns with recent prototype-based personalization methods \cite{liu2025model}. 

To further examine the sensitivity of prototype sharing, we evaluate the robustness of FRAMP under two forms of prototype perturbation. Gaussian noise and random rotation are commonly used perturbations, and here they serve as stress tests to assess how strongly the model depends on the exact prototype values. 
\cref{tab:protopertub} shows representative results under strong perturbations (GN $a=0.5$ and random rotation), where degradation occurs. More results across submodel sizes are in \Cref{app:privacyanal}, which overall show minimal accuracy loss in most cases, suggesting that FRAMP leverages prototypes mainly for semantic guidance rather than precise reconstruction.

Although FRAMP involves uploading masked model updates, its use of personalized submodels and dynamic masking limits the amount of consistent gradient information observed by the server. These mechanisms reduce cross-round consistency that gradient inversion attacks typically rely on. Moreover, FRAMP remains compatible with standard privacy-enhancing techniques, such as secure aggregation and differential privacy, which can be applied in parallel to provide stronger privacy protection.

% %%%%%%%%%%%%%%%%%%
\begin{table}[!t]
\caption{Server training time per communication round on CIFAR-10 (in milliseconds).}
\label{tab:runtime}
\vspace{-1mm}
\centering
\resizebox{\linewidth}{!}{
\begin{tabular}{lccccc}
\toprule
\textbf{Method} & HeteroFL & FedRolex & ScaleFL & FIARSE & FRAMP \\
\midrule
\textbf{Rnd 1}  & 24.74  & 54.05  & 25.45  & 26.89  & 37.96 \\
\textbf{Rnd 2+} & 8.01   & 8.09   & 8.35   & 7.67   & 7.98  \\
\bottomrule
\end{tabular}}
\vspace{-3mm}
\end{table}
\begin{table}[!t]
\caption{Test accuracy under Gaussian noise and random rotation on CIFAR-100.}
\label{tab:protopertub}
\vspace{-1mm}
\centering
\resizebox{0.9\linewidth}{!}{
\begin{tabular}{lccccc}
\toprule
\textbf{} & \shortstack{No\\Noise} & \shortstack{GN \\ $a=0.01$} & \shortstack{GN \\ $a=0.1$} & \shortstack{GN \\ $a=0.5$} & \shortstack{Random\\Rotation} \\
\midrule
\textbf{Local}  & 42.95  & 42.40  & 41.45  & 41.25  & 41.73 \\
\textbf{Union}  & 40.26  & 40.26  & 40.04  & 38.41  & 39.02 \\
\bottomrule
\end{tabular}}
\vspace{-5mm}
\end{table}
%%%%%%%%%%%%%%%%%%%%%%%%%%

%%%%%%%%%%%%%%%%%%%%%%%%%%%%%%%%%%%%
\textbf{Additional Experiments in Appendix}\hspace{0.7em}
\Cref{app:5modelsize} reports results under severe system heterogeneity with five model sizes. 
\Cref{app:diffalpha} presents experiments under different levels of data heterogeneity, including stronger heterogeneity ($\alpha=0.1, 0.05$), milder heterogeneity ($\alpha=0.5, 0.7$), and the IID setting.
\Cref{app:20paticipants} evaluates scenarios with more participants per round.

%%%%%%%%%%%%%%%%%%%%%%%%%%%%%%%%%%%%%%%%%%%%%%%%%%%%%%%%%%%%%%%
\vspace{-1mm}
\section{Conclusion}

We proposed FRAMP, a unified framework for personalized and resource-adaptive FL. By integrating client-aware model generation, adaptive submodel extraction, and semantic alignment, FRAMP enables clients to obtain effective and personalized submodels. 
Extensive experiments demonstrate robust personalization and generalization. FRAMP achieves more balanced performance across model-size groups, maintains strong accuracy for clients with limited budgets, and generalizes well to unseen compute budgets and new clients. 
While our study focuses on magnitude-based importance and hypernetwork-based model generation, exploring alternative importance estimators or more expressive generators is a promising future direction.

\bibliographystyle{ICML2026/icml2026}
\bibliography{arxiv_FRAMP_main}

@inproceedings{baek2023personalized,
  title={Personalized subgraph federated learning},
  author={Baek, Jinheon and Jeong, Wonyong and Jin, Jiongdao and Yoon, Jaehong and Hwang, Sung Ju},
  booktitle={International Conference on Machine Learning},
  pages={1396--1415},
  year={2023},
  organization={PMLR}
}

@article{li2020federated,
  title={Federated optimization in heterogeneous networks},
  author={Li, Tian and Sahu, Anit Kumar and Zaheer, Manzil and Sanjabi, Maziar and Talwalkar, Ameet and Smith, Virginia},
  journal={Proceedings of Machine learning and systems},
  volume={2},
  pages={429--450},
  year={2020}
}

@article{arivazhagan2019federated,
  title={Federated learning with personalization layers},
  author={Arivazhagan, Manoj Ghuhan and Aggarwal, Vinay and Singh, Aaditya Kumar and Choudhary, Sunav},
  journal={arXiv preprint arXiv:1912.00818},
  year={2019}
}

@article{lin2020ensemble,
  title={Ensemble distillation for robust model fusion in federated learning},
  author={Lin, Tao and Kong, Lingjing and Stich, Sebastian U and Jaggi, Martin},
  journal={Advances in Neural Information Processing Systems},
  volume={33},
  pages={2351--2363},
  year={2020}
}

@inproceedings{klein2015dynamic,
  title={A dynamic convolutional layer for short range weather prediction},
  author={Klein, Benjamin and Wolf, Lior and Afek, Yehuda},
  booktitle={Proceedings of the IEEE Conference on Computer Vision and Pattern Recognition},
  pages={4840--4848},
  year={2015}
}

@inproceedings{klocek2019hypernetwork,
  title={Hypernetwork functional image representation},
  author={Klocek, Sylwester and Maziarka, {\L}ukasz and Wo{\l}czyk, Maciej and Tabor, Jacek and Nowak, Jakub and {\'S}mieja, Marek},
  booktitle={International Conference on Artificial Neural Networks},
  pages={496--510},
  year={2019},
  organization={Springer}
}

@article{navon2020learning,
  title={Learning the pareto front with hypernetworks},
  author={Navon, Aviv and Shamsian, Aviv and Chechik, Gal and Fetaya, Ethan},
  journal={arXiv preprint arXiv:2010.04104},
  year={2020}
}

@article{brock2017smash,
  title={{SMASH}: one-shot model architecture search through hypernetworks},
  author={Brock, Andrew and Lim, Theodore and Ritchie, James M and Weston, Nick},
  journal={arXiv preprint arXiv:1708.05344},
  year={2017}
}

@inproceedings{xu2023heterogeneous,
  title={Heterogeneous Federated Learning Based on Graph Hypernetwork},
  author={Xu, Zhengyi and Yang, Liu and Gu, Shiqiao},
  booktitle={International Conference on Artificial Neural Networks},
  pages={464--476},
  year={2023},
  organization={Springer}
}

@inproceedings{ye2023personalized,
  title={Personalized federated learning with inferred collaboration graphs},
  author={Ye, Rui and Ni, Zhenyang and Wu, Fangzhao and Chen, Siheng and Wang, Yanfeng},
  booktitle={International Conference on Machine Learning},
  pages={39801--39817},
  year={2023},
  organization={PMLR}
}

@article{zhang2018graph,
  title={Graph hypernetworks for neural architecture search},
  author={Zhang, Chris and Ren, Mengye and Urtasun, Raquel},
  journal={arXiv preprint arXiv:1810.05749},
  year={2018}
}

@inproceedings{shamsian2021personalized,
  title={Personalized federated learning using hypernetworks},
  author={Shamsian, Aviv and Navon, Aviv and Fetaya, Ethan and Chechik, Gal},
  booktitle={International Conference on Machine Learning},
  pages={9489--9502},
  year={2021},
  organization={PMLR}
}

@article{hu2020open,
  title={Open graph benchmark: Datasets for machine learning on graphs},
  author={Hu, Weihua and Fey, Matthias and Zitnik, Marinka and Dong, Yuxiao and Ren, Hongyu and Liu, Bowen and Catasta, Michele and Leskovec, Jure},
  journal={Advances in neural information processing systems},
  volume={33},
  pages={22118--22133},
  year={2020}
}

@article{karypis1997metis,
  title={METIS: Unstructured graph partitioning and sparse matrix ordering system},
  author={Karypis, George},
  journal={Technical report},
  year={1997},
  publisher={Department of Computer Science, University of Minnesota}
}

@inproceedings{schneider2021personalization,
  title={Personalization of deep learning},
  author={Schneider, Johannes and Vlachos, Michalis},
  booktitle={Proceedings of International Data Science Conference},
  pages={89--96},
  year={2021},
  organization={Springer}
}

@article{hanzely2020lower,
  title={Lower bounds and optimal algorithms for personalized federated learning},
  author={Hanzely, Filip and Hanzely, Slavom{\'\i}r and Horv{\'a}th, Samuel and Richt{\'a}rik, Peter},
  journal={Advances in Neural Information Processing Systems},
  volume={33},
  pages={2304--2315},
  year={2020}
}

@inproceedings{ma2022layer,
  title={Layer-wised model aggregation for personalized federated learning},
  author={Ma, Xiaosong and Zhang, Jie and Guo, Song and Xu, Wenchao},
  booktitle={Proceedings of the IEEE/CVF conference on computer vision and pattern recognition},
  pages={10092--10101},
  year={2022}
}

@article{jiang2019improving,
  title={Improving federated learning personalization via model agnostic meta learning},
  author={Jiang, Yihan and Kone{\v{c}}n{\`y}, Jakub and Rush, Keith and Kannan, Sreeram},
  journal={arXiv preprint arXiv:1909.12488},
  year={2019}
}

@article{lee2024fedl2p,
  title={FedL2P: Federated Learning to Personalize},
  author={Lee, Royson and Kim, Minyoung and Li, Da and Qiu, Xinchi and Hospedales, Timothy and Husz{\'a}r, Ferenc and Lane, Nicholas},
  journal={Advances in Neural Information Processing Systems},
  volume={36},
  year={2024}
}

@article{ghosh2020efficient,
  title={An efficient framework for clustered federated learning},
  author={Ghosh, Avishek and Chung, Jichan and Yin, Dong and Ramchandran, Kannan},
  journal={Advances in Neural Information Processing Systems},
  volume={33},
  pages={19586--19597},
  year={2020}
}

@inproceedings{huang2021personalized,
  title={Personalized cross-silo federated learning on non-iid data},
  author={Huang, Yutao and Chu, Lingyang and Zhou, Zirui and Wang, Lanjun and Liu, Jiangchuan and Pei, Jian and Zhang, Yong},
  booktitle={Proceedings of the AAAI conference on Artificial Intelligence},
  pages={7865--7873},
  year={2021}
}

@article{zhang2021parameterized,
  title={Parameterized knowledge transfer for personalized federated learning},
  author={Zhang, Jie and Guo, Song and Ma, Xiaosong and Wang, Haozhao and Xu, Wenchao and Wu, Feijie},
  journal={Advances in Neural Information Processing Systems},
  volume={34},
  pages={10092--10104},
  year={2021}
}

@article{yan2024personalized,
  title={Personalized Federated Learning With Multi-View Geometry Structure},
  author={Yan, Yihan and Wang, Shen and Sun, Fanghui and Tong, Xiaojun},
  journal={IEEE Internet of Things Journal},
  year={2024},
  publisher={IEEE}
}

@article{konevcny2016federated,
  title={Federated Learning: Strategies for Improving Communication Efficiency},
  author={Kone{\v{c}}n{\`y}, Jakub},
  journal={arXiv preprint arXiv:1610.05492},
  year={2016}
}

@inproceedings{scott2024pefll,
  title={PeFLL: Personalized federated learning by learning to learn},
  author={Scott, Jonathan A and Zakerinia, Hossein and Lampert, Christoph},
  booktitle={12th International Conference on Learning Representations},
  year={2024}
}

@article{wu2024fiarse,
  title={FIARSE: Model-heterogeneous federated learning via importance-aware submodel extraction},
  author={Wu, Feijie and Wang, Xingchen and Wang, Yaqing and Liu, Tianci and Su, Lu and Gao, Jing},
  journal={Advances in Neural Information Processing Systems},
  volume={37},
  pages={115615--115651},
  year={2024}
}

@article{jayakumar2020top,
  title={Top-kast: Top-k always sparse training},
  author={Jayakumar, Siddhant and Pascanu, Razvan and Rae, Jack and Osindero, Simon and Elsen, Erich},
  journal={Advances in Neural Information Processing Systems},
  volume={33},
  pages={20744--20754},
  year={2020}
}

@inproceedings{mostafa2019parameter,
  title={Parameter efficient training of deep convolutional neural networks by dynamic sparse reparameterization},
  author={Mostafa, Hesham and Wang, Xin},
  booktitle={International Conference on Machine Learning},
  pages={4646--4655},
  year={2019},
  organization={PMLR}
}

@article{mitra2021linear,
  title={Linear convergence in federated learning: Tackling client heterogeneity and sparse gradients},
  author={Mitra, Aritra and Jaafar, Rayana and Pappas, George J and Hassani, Hamed},
  journal={Advances in Neural Information Processing Systems},
  volume={34},
  pages={14606--14619},
  year={2021}
}

@inproceedings{shin2022fedbalancer,
  title={Fedbalancer: Data and pace control for efficient federated learning on heterogeneous clients},
  author={Shin, Jaemin and Li, Yuanchun and Liu, Yunxin and Lee, Sung-Ju},
  booktitle={Proceedings of the 20th Annual International Conference on Mobile Systems, Applications and Services},
  pages={436--449},
  year={2022}
}

@article{wu2023deterioration,
  title={From deterioration to acceleration: A calibration approach to rehabilitating step asynchronism in federated optimization},
  author={Wu, Feijie and Guo, Song and Wang, Haozhao and Zhang, Haobo and Qu, Zhihao and Zhang, Jie and Liu, Ziming},
  journal={IEEE Transactions on Parallel and Distributed Systems},
  volume={34},
  number={5},
  pages={1548--1559},
  year={2023},
  publisher={IEEE}
}

@article{cho2022heterogeneous,
  title={Heterogeneous ensemble knowledge transfer for training large models in federated learning},
  author={Cho, Yae Jee and Manoel, Andre and Joshi, Gauri and Sim, Robert and Dimitriadis, Dimitrios},
  journal={arXiv preprint arXiv:2204.12703},
  year={2022}
}

@article{zhang2023distill,
  title={To distill or not to distill: Toward fast, accurate, and communication-efficient federated distillation learning},
  author={Zhang, Yuan and Zhang, Wenlong and Pu, Lingjun and Lin, Tao and Yan, Jinyao},
  journal={IEEE Internet of Things Journal},
  volume={11},
  number={6},
  pages={10040--10053},
  year={2023},
  publisher={IEEE}
}

@inproceedings{alballa2023first,
  title={A first look at the impact of distillation hyper-parameters in federated knowledge distillation},
  author={Alballa, Norah and Canini, Marco},
  booktitle={Proceedings of the 3rd Workshop on Machine Learning and Systems},
  pages={123--130},
  year={2023}
}

@inproceedings{chen2023efficient,
  title={Efficient personalized federated learning via sparse model-adaptation},
  author={Chen, Daoyuan and Yao, Liuyi and Gao, Dawei and Ding, Bolin and Li, Yaliang},
  booktitle={International conference on machine learning},
  pages={5234--5256},
  year={2023},
  organization={PMLR}
}

@inproceedings{liao2023adaptive,
  title={Adaptive channel sparsity for federated learning under system heterogeneity},
  author={Liao, Dongping and Gao, Xitong and Zhao, Yiren and Xu, Cheng-Zhong},
  booktitle={Proceedings of the IEEE/CVF Conference on Computer Vision and Pattern Recognition},
  pages={20432--20441},
  year={2023}
}

@article{zhou2023every,
  title={Every parameter matters: Ensuring the convergence of federated learning with dynamic heterogeneous models reduction},
  author={Zhou, Hanhan and Lan, Tian and Venkataramani, Guru Prasadh and Ding, Wenbo},
  journal={Advances in Neural Information Processing Systems},
  volume={36},
  pages={25991--26002},
  year={2023}
}

@article{diao2020heterofl,
  title={Heterofl: Computation and communication efficient federated learning for heterogeneous clients},
  author={Diao, Enmao and Ding, Jie and Tarokh, Vahid},
  journal={arXiv preprint arXiv:2010.01264},
  year={2020}
}

@inproceedings{ilhan2023scalefl,
  title={Scalefl: Resource-adaptive federated learning with heterogeneous clients},
  author={Ilhan, Fatih and Su, Gong and Liu, Ling},
  booktitle={Proceedings of the IEEE/CVF Conference on Computer Vision and Pattern Recognition},
  pages={24532--24541},
  year={2023}
}

@article{kang2023nefl,
  title={Nefl: Nested model scaling for federated learning with system heterogeneous clients},
  author={Kang, Honggu and Cha, Seohyeon and Shin, Jinwoo and Lee, Jongmyeong and Kang, Joonhyuk},
  journal={arXiv preprint arXiv:2308.07761},
  year={2023}
}

@inproceedings{kim2022depthfl,
  title={Depthfl: Depthwise federated learning for heterogeneous clients},
  author={Kim, Minjae and Yu, Sangyoon and Kim, Suhyun and Moon, Soo-Mook},
  booktitle={The Eleventh International Conference on Learning Representations},
  year={2022}
}

@inproceedings{wang2023flexifed,
  title={Flexifed: Personalized federated learning for edge clients with heterogeneous model architectures},
  author={Wang, Kaibin and He, Qiang and Chen, Feifei and Chen, Chunyang and Huang, Faliang and Jin, Hai and Yang, Yun},
  booktitle={Proceedings of the ACM Web Conference 2023},
  pages={2979--2990},
  year={2023}
}

@article{yi2024fedp3,
  title={Fedp3: Federated personalized and privacy-friendly network pruning under model heterogeneity},
  author={Yi, Kai and Gazagnadou, Nidham and Richt{\'a}rik, Peter and Lyu, Lingjuan},
  journal={arXiv preprint arXiv:2404.09816},
  year={2024}
}

@article{alam2022fedrolex,
  title={Fedrolex: Model-heterogeneous federated learning with rolling sub-model extraction},
  author={Alam, Samiul and Liu, Luyang and Yan, Ming and Zhang, Mi},
  journal={Advances in neural information processing systems},
  volume={35},
  pages={29677--29690},
  year={2022}
}

@article{caldas2018expanding,
  title={Expanding the reach of federated learning by reducing client resource requirements},
  author={Caldas, Sebastian and Kone{\v{c}}ny, Jakub and McMahan, H Brendan and Talwalkar, Ameet},
  journal={arXiv preprint arXiv:1812.07210},
  year={2018}
}

@article{horvath2021fjord,
  title={Fjord: Fair and accurate federated learning under heterogeneous targets with ordered dropout},
  author={Horvath, Samuel and Laskaridis, Stefanos and Almeida, Mario and Leontiadis, Ilias and Venieris, Stylianos and Lane, Nicholas},
  journal={Advances in Neural Information Processing Systems},
  volume={34},
  pages={12876--12889},
  year={2021}
}

@article{liang2025personalized,
  title={Personalized Subgraph Federated Learning with Sheaf Collaboration},
  author={Liang, Wenfei and Zhao, Yanan and She, Rui and Li, Yiming and Tay, Wee Peng},
  journal={arXiv preprint arXiv:2508.13642},
  year={2025}
}

@article{krizhevsky2009learning,
  title={Learning multiple layers of features from tiny images},
  author={Krizhevsky, Alex and Hinton, Geoffrey and others},
  year={2009},
  publisher={Toronto, ON, Canada}
}

@article{frankle2018lottery,
  title={The lottery ticket hypothesis: Finding sparse, trainable neural networks},
  author={Frankle, Jonathan and Carbin, Michael},
  journal={arXiv preprint arXiv:1803.03635},
  year={2018}
}

@inproceedings{liu2015sparse,
  title={Sparse convolutional neural networks},
  author={Liu, Baoyuan and Wang, Min and Foroosh, Hassan and Tappen, Marshall and Pensky, Marianna},
  booktitle={Proceedings of the IEEE conference on computer vision and pattern recognition},
  pages={806--814},
  year={2015}
}

@article{hoefler2021sparsity,
  title={Sparsity in deep learning: Pruning and growth for efficient inference and training in neural networks},
  author={Hoefler, Torsten and Alistarh, Dan and Ben-Nun, Tal and Dryden, Nikoli and Peste, Alexandra},
  journal={Journal of Machine Learning Research},
  volume={22},
  number={241},
  pages={1--124},
  year={2021}
}

@inproceedings{iofinova2022well,
  title={How well do sparse imagenet models transfer?},
  author={Iofinova, Eugenia and Peste, Alexandra and Kurtz, Mark and Alistarh, Dan},
  booktitle={Proceedings of the IEEE/CVF Conference on Computer Vision and Pattern Recognition},
  pages={12266--12276},
  year={2022}
}

@inproceedings{kurtz2020inducing,
  title={Inducing and exploiting activation sparsity for fast inference on deep neural networks},
  author={Kurtz, Mark and Kopinsky, Justin and Gelashvili, Rati and Matveev, Alexander and Carr, John and Goin, Michael and Leiserson, William and Moore, Sage and Shavit, Nir and Alistarh, Dan},
  booktitle={International Conference on Machine Learning},
  pages={5533--5543},
  year={2020},
  organization={PMLR}
}

@article{isik2022sparse,
  title={Sparse random networks for communication-efficient federated learning},
  author={Isik, Berivan and Pase, Francesco and Gunduz, Deniz and Weissman, Tsachy and Zorzi, Michele},
  journal={arXiv preprint arXiv:2209.15328},
  year={2022}
}

@article{jiang2022model,
  title={Model pruning enables efficient federated learning on edge devices},
  author={Jiang, Yuang and Wang, Shiqiang and Valls, Victor and Ko, Bong Jun and Lee, Wei-Han and Leung, Kin K and Tassiulas, Leandros},
  journal={IEEE Transactions on Neural Networks and Learning Systems},
  volume={34},
  number={12},
  pages={10374--10386},
  year={2022},
  publisher={IEEE}
}

@inproceedings{li2021fedmask,
  title={Fedmask: Joint computation and communication-efficient personalized federated learning via heterogeneous masking},
  author={Li, Ang and Sun, Jingwei and Zeng, Xiao and Zhang, Mi and Li, Hai and Chen, Yiran},
  booktitle={Proceedings of the 19th ACM conference on embedded networked sensor systems},
  pages={42--55},
  year={2021}
}

@inproceedings{bibikar2022federated,
  title={Federated dynamic sparse training: Computing less, communicating less, yet learning better},
  author={Bibikar, Sameer and Vikalo, Haris and Wang, Zhangyang and Chen, Xiaohan},
  booktitle={Proceedings of the AAAI Conference on Artificial Intelligence},
  volume={36},
  number={6},
  pages={6080--6088},
  year={2022}
}

@article{dai2022dispfl,
  title={Dispfl: Towards communication-efficient personalized federated learning via decentralized sparse training},
  author={Dai, Rong and Shen, Li and He, Fengxiang and Tian, Xinmei and Tao, Dacheng},
  journal={arXiv preprint arXiv:2206.00187},
  year={2022}
}

@article{li2020lotteryfl,
  title={Lotteryfl: Personalized and communication-efficient federated learning with lottery ticket hypothesis on non-iid datasets},
  author={Li, Ang and Sun, Jingwei and Wang, Binghui and Duan, Lin and Li, Sicheng and Chen, Yiran and Li, Hai},
  journal={arXiv preprint arXiv:2008.03371},
  year={2020}
}

@inproceedings{mugunthan2022fedltn,
  title={Fedltn: Federated learning for sparse and personalized lottery ticket networks},
  author={Mugunthan, Vaikkunth and Lin, Eric and Gokul, Vignesh and Lau, Christian and Kagal, Lalana and Pieper, Steve},
  booktitle={European Conference on Computer Vision},
  pages={69--85},
  year={2022},
  organization={Springer}
}

@inproceedings{seo2021communication,
  title={Communication-efficient and personalized federated lottery ticket learning},
  author={Seo, Sejin and Ko, Seung-Woo and Park, Jihong and Kim, Seong-Lyun and Bennis, Mehdi},
  booktitle={2021 IEEE 22nd International Workshop on Signal Processing Advances in Wireless Communications (SPAWC)},
  pages={581--585},
  year={2021},
  organization={IEEE}
}

@ARTICLE{liu2025model,
  author={Liu, Zhi and Zhou, Hanlin and He, Xiaohua and Yuan, Haopeng and Du, Jiaxin and Wang, Mengmeng and Shen, Guojiang and Kong, Xiangjie and Xia, Feng},
  journal={IEEE Transactions on Artificial Intelligence}, 
  title={Model-Heterogeneous Federated Graph Learning With Prototype Propagation}, 
  year={2025},
  volume={6},
  number={3},
  pages={676-689},
  doi={10.1109/TAI.2024.3490557}}

@inproceedings{wang2024feddse,
  title={Feddse: Distribution-aware sub-model extraction for federated learning over resource-constrained devices},
  author={Wang, Haozhao and Jia, Yabo and Zhang, Meng and Hu, Qinghao and Ren, Hao and Sun, Peng and Wen, Yonggang and Zhang, Tianwei},
  booktitle={Proceedings of the ACM Web Conference 2024},
  pages={2902--2913},
  year={2024}
}

@inproceedings{deng2022tailorfl,
  title={TailorFL: Dual-personalized federated learning under system and data heterogeneity},
  author={Deng, Yongheng and Chen, Weining and Ren, Ju and Lyu, Feng and Liu, Yang and Liu, Yunxin and Zhang, Yaoxue},
  booktitle={Proceedings of the 20th ACM conference on embedded networked sensor systems},
  pages={592--606},
  year={2022}
}

@article{chen2025advances,
  title={Advances in robust federated learning: A survey with heterogeneity considerations},
  author={Chen, Chuan and Liao, Tianchi and Deng, Xiaojun and Wu, Zihou and Huang, Sheng and Zheng, Zibin},
  journal={IEEE Transactions on Big Data},
  year={2025},
  publisher={IEEE}
}

@inproceedings{zhang2025htfllib,
  title={Htfllib: A comprehensive heterogeneous federated learning library and benchmark},
  author={Zhang, Jianqing and Wu, Xinghao and Zhou, Yanbing and Sun, Xiaoting and Cai, Qiqi and Liu, Yang and Hua, Yang and Zheng, Zhenzhe and Cao, Jian and Yang, Qiang},
  booktitle={Proceedings of the 31st ACM SIGKDD Conference on Knowledge Discovery and Data Mining V. 2},
  pages={5900--5911},
  year={2025}
}

@article{kim2025subgraph,
  title={Subgraph federated learning for local generalization},
  author={Kim, Sungwon and Lee, Yoonho and Oh, Yunhak and Lee, Namkyeong and Yun, Sukwon and Lee, Junseok and Kim, Sein and Yang, Carl and Park, Chanyoung},
  journal={arXiv preprint arXiv:2503.03995},
  year={2025}
}

%%%%%%%%%%%%%%%%%%%%%%%%%%%%%%%%%%%%%%%%%%%%%%%%%%%%%%%%%%%%%%%%%%%%%%%%%%%%%%%
%%%%%%%%%%%%%%%%%%%%%%%%%%%%%%%%%%%%%%%%%%%%%%%%%%%%%%%%%%%%%%%%%%%%%%%%%%%%%%%
% APPENDIX
%%%%%%%%%%%%%%%%%%%%%%%%%%%%%%%%%%%%%%%%%%%%%%%%%%%%%%%%%%%%%%%%%%%%%%%%%%%%%%%
%%%%%%%%%%%%%%%%%%%%%%%%%%%%%%%%%%%%%%%%%%%%%%%%%%%%%%%%%%%%%%%%%%%%%%%%%%%%%%%
\newpage
\appendix
\onecolumn

\section{Related Works}\label{app:relatedwork}
\subsection{Model Sparsification in FL}
Model sparsification, or pruning, has gained attention following the introduction of the lottery ticket hypothesis \cite{frankle2018lottery}, which suggests that within large models lie smaller subnetworks that can be trained to perform effectively. Early studies \cite{liu2015sparse} leveraged sparsity to reduce computational overhead. Recently, hardware innovations have further advanced the feasibility of training and deploying sparse models \cite{kurtz2020inducing,hoefler2021sparsity,iofinova2022well}.

In \gls{FL}, two main approaches are used to obtain sparse submodels: dense-to-sparse \cite{li2021fedmask,isik2022sparse,deng2022tailorfl,jiang2022model,wang2024feddse}, which begins with a full model, and sparse-to-sparse \cite{li2020lotteryfl,seo2021communication,bibikar2022federated,dai2022dispfl,mugunthan2022fedltn}, which starts with a sparse model. Both methods typically distribute a shared global model to clients, and then prune or refine it locally. However, these techniques struggle under heterogeneous resource constraints, where some clients cannot even load the full model due to limited capacity.

To address this, algorithms have emerged that generate submodels tailored to each client’s computational budget. For example, Flado \cite{liao2023adaptive} allows the server to customize submodels based on client computation budgets by ranking parameters according to importance. While effective, this method requires clients to store and update importance scores for all parameters, introducing non-trivial storage and computation overhead.
FIARSE \cite{wu2024fiarse} avoids this burden by leveraging the observation that parameter magnitude often correlates with importance. It uses parameter values themselves as implicit importance indicators, thereby simplifying the process and reducing overhead on resource-limited clients.

\subsection{Hypernetworks}
\Glspl{HN} \cite{klein2015dynamic} are neural networks designed to generate weights for another network. The target weights can be dynamically adapted based on the HNs' input vectors. \cite{klocek2019hypernetwork,navon2020learning}.
SMASH \cite{brock2017smash} introduced HNs into Neural Architecture Search (NAS) by encoding architectures as 3D tensors using a memory channel mechanism. 
GHN \cite{zhang2018graph} was proposed for anytime prediction, focusing on both inference efficiency and the speed-accuracy trade-off. In FL settings, pFedHN \cite{shamsian2021personalized} employed HNs with task-specific embeddings for personalization, while hFedGHN \cite{xu2023heterogeneous} extended GHNs to support heterogeneous local models through graph-based reasoning. 

Direct comparison with HN-based FL baselines is not applicable, as they are designed for uniform models and do not support varying model sizes.

%%%%%%%%%%%%%%%%%%%%%%
\section{Algorithm}\label{app:algorithm}
\cref{alg:framp} details the full procedure of FRAMP.

\begin{algorithm}[H]
\caption{FRAMP}
\label{alg:framp}
\noindent\textbf{Input:} Communication rounds $R$, number of clients $N$, local training rounds $T$.

\begin{minipage}[t]{0.5\textwidth}
\noindent\textbf{Server:}
\begin{algorithmic}[1]
\FOR{$r = 1$ to $R$}
    \STATE Sample a subset of clients $\calA\subseteq[N]$.
    \FORALL{$n \in \calA$}
        \STATE Generate customized full model $\bomega_n^r := H(\bv_n;\bm{\varphi}^r)$.
        \STATE Compute client-specific mask $\calM_n^r(\bomega_n^r)$.
        \STATE Send submodel $\hat\bomega_n^r=\bomega_n^r\odot\calM_n^r(\bomega_n^r)$ and global prototypes $\bP_g^r$ to client $n$.
        \STATE \textbf{ClientUpdate}($n$, $\hat\bomega_n^r$, $\bP_g^r$)
    \ENDFOR
    \STATE Update $\bm{\varphi}^{r+1}=\bm{\varphi}^r-\beta\left( \nabla_{\bm{\varphi}^r} \hat\bomega_n^r \right)^\top \Delta \hat\bomega_n^r$.
    \STATE Update $\bP_g^{r+1}\gets\left\{\bP_n^{r,c}\right\}_{c=1}^C$.
\ENDFOR
\end{algorithmic}
\end{minipage}
\hfill
\begin{minipage}[t]{0.49\textwidth}
\noindent\textbf{ClientUpdate ($n$, $\hat\bomega_n^r$, $\bP_g^r$):}

\begin{algorithmic}[1]
\STATE Set $\hat\bomega_n^{r,1} = \hat\bomega_n^r$.
\FOR{$t = 1$ to $T$}
    \STATE Update parameters 
    $\hat\bomega_n^{r,t+1} = \hat\bomega_n^{r,t} - \alpha \nabla_{\hat\bomega_n^{r,t}} \mathcal{L}_n(\hat\bomega_n^{r,t})$.
    \STATE Update local prototypes $\left\{\bP_n^{r,c}\right\}_{c=1}^C$.
\ENDFOR
\STATE Compute update: $\Delta \hat\bomega_n^r = \hat\bomega_n^{r,T} - \hat\bomega_n^{r,1}$.
\STATE \textbf{return} $\Delta \hat\bomega_n^r$, $\left\{\bP_n^{r,c}\right\}_{c=1}^C$
\end{algorithmic}
\end{minipage}
\end{algorithm}

%%%%%%%%%%%%%%%%%%%%%%%%%%%%%%%%
\section{Experimental Setups}\label{app:setup}
\subsection{Datasets}\label{app:dataset}
For CIFAR-10 and CIFAR-100, the datasets contain 50K training samples and 10K test samples, which are then divided among 100 clients using a Dirichlet distribution with concentration parameter $\alpha = 0.3$.
For ogbn-arxiv, 40\% of nodes are used for training, 30\% for validation, and the remaining 30\% for testing. To distribute the graph among clients, we employ the METIS graph partitioning algorithm \cite{karypis1997metis}, which segments the original graph into a specified number of disjoint subgraphs. Each client is assigned one of these non-overlapping subgraphs. In our case, setting METIS to 100 yields 100 disjoint subgraphs, each corresponding to a client in the federated setting.

\subsection{Hyperparameters}\label{app:hyperparameters}
\cref{tab:hyperparams} lists the hyperparameters used in the experiments.

\begin{table}[h]
\centering
\caption{Hyperparameter Settings}
\label{tab:hyperparams}
\resizebox{0.55\linewidth}{!}{
\begin{tabular}{lccc}
\toprule
 & \textbf{CIFAR-10} & \textbf{CIFAR-100} & \textbf{ogbn-arxiv} \\
\midrule
Local Epochs           & 100   & 100   & 120   \\
Batch Size             & 32    & 32    & -  \\
Communication Rounds   & 800 & 800 & 200 \\
Learning rate          & \{0.1, 0.5\} & \{0.1, 0.4\} & \{0.1\} \\
HN Learning rate       & 0.12  &0.08  & 0.1 \\   
$\lambda$              & 0.7   &0.7   & 0.2 \\
\bottomrule
\end{tabular}
}
\end{table}

%%%%%%%%%%%%%%%%%%%%%%%%%%%%%%%%
\section{More Experiments}\label{app:moreexp}

%%%%%%%%%%%%%%%%%%%%%%%%%%%%%%%%
\subsection{System Heterogeneity}\label{app:5modelsize}
\textbf{System heterogeneity with five model sizes}\hspace{0.7em}
We evaluate a setting with five model capacities, $\gamma' = {0.04, 0.16, 0.36, 0.64, 1.0}$, with clients evenly assigned to each level. This experiment illustrates the flexibility of our approach, which can naturally extend to different numbers of capacity levels or alternative client distributions.

\begin{table}[h]
\caption{Test accuracy of submodels across five submodel sizes.}
\label{tb:5modelacc}
\centering
\resizebox{0.6\textwidth}{!}{
\begin{tabular}{lccccccc}
\toprule
\multirow{2}{*}{Method} & \multicolumn{7}{c}{CIFAR-100} \\
\cmidrule(lr){2-8}
&Local & 0.04 & 0.16 & 0.36 & 0.64 & 1.0 &Union\\
\midrule
HeteroFL &35.01  &30.15  &33.60  &36.10  &37.75  &37.45  &32.60 \\
FedRolex &38.51  &28.40  &37.75  &41.10  &42.10  &43.19  &36.09 \\
ScaleFL  &42.43  &42.90  &44.75  &42.49  &42.05  &39.95  &40.21 \\
FIARSE   &45.94  &44.35  &45.65  &47.80  &47.00  &44.90  &42.61 \\
FRAMP    &\textbf{46.71} &\textbf{46.75} &\textbf{46.65} &\textbf{47.85} &\textbf{47.35} &\textbf{44.95} &\textbf{43.67}\\
\bottomrule
\end{tabular}
}
\end{table}

%%%%%%%%%%%%%%%%%%%%%%%%%%%%%%%%
\subsection{Data Heterogeneity}\label{app:diffalpha}
To examine the effect of stronger data heterogeneity, we partition CIFAR-100 into 100 clients using a Dirichlet distribution with concentration parameter $\alpha=0.1$ and $\alpha=0.05$. The results are reported in \cref{tb:alpha0.1} and \cref{tb:alpha0.05}. Compared with the default $\alpha=0.3$, this setting yields a more skewed label distribution across clients, providing a stricter test of robustness under highly non-IID conditions.

To explore the impact of less heterogeneous setting, we conduct experiments with $\alpha=0.5$ and $\alpha=0.7$, where client data distributions become more balanced. The corresponding results are presented in \cref{tb:alpha0.5} and \cref{tb:alpha0.7}.

In addition, we include experiments under the IID partition with results shown in \cref{tb:iid}.

%%%%%%%%%%%%%%%%%%%%%%%%%%%%%%%%
\begin{table}[!h]
\caption{Test accuracy on CIFAR-100 with stronger data heterogeneity ($\alpha=0.05$).}
\label{tb:alpha0.05}
\vspace{-0.5mm}
\centering
\resizebox{0.55\textwidth}{!}{
\begin{tabular}{lcccccc}
\toprule
\multirow{2}{*}{Method} & \multicolumn{6}{c}{CIFAR-100}\\
\cmidrule(lr){2-7}
&Local & 1/64 & 1/16 & 1/4 & 1.0 &Union\\
\midrule
HeteroFL &22.24  &17.28  &21.96  &22.16  &27.56  &21.48     \\
FedRolex &12.63  &3.08  &5.84    &16.12  &25.48  &11.67    \\
ScaleFL  &29.14  &28.68  &32.04  &28.64  &27.20  &27.71    \\
FIARSE   &29.29  &25.68  &31.84  &\textbf{29.24}  &30.40  &29.25     \\
FRAMP    &\textbf{30.92}  &\textbf{31.32}  &\textbf{33.12}  &28.72  &\textbf{30.52}  &\textbf{30.10} \\
\bottomrule
\end{tabular}
}
\end{table}
%%%%%%%%%%%%%%%%%%%%%%%%%%%%%%%%
\begin{table}[!h]
\caption{Test accuracy on CIFAR-100 with stronger data heterogeneity ($\alpha=0.1$).}
\label{tb:alpha0.1}
\vspace{-0.5mm}
\centering
\resizebox{0.55\textwidth}{!}{
\begin{tabular}{lcccccc}
\toprule
\multirow{2}{*}{Method} & \multicolumn{6}{c}{CIFAR-100}\\
\cmidrule(lr){2-7}
&Local & 1/64 & 1/16 & 1/4 & 1.0 &Union\\
\midrule
HeteroFL &25.80  &20.68  &23.92  &29.36  &29.24  &23.79     \\
FedRolex &16.06  &2.56   &8.12   &23.32  &30.24  &15.04     \\
ScaleFL  &33.74  &31.36  &35.08  &37.76  &30.76  &31.61     \\
FIARSE   &34.64  &29.92  &36.04  &\textbf{40.08}  &32.52  &34.10     \\
FRAMP    &\textbf{36.57} &\textbf{36.56} &\textbf{38.16} &38.96 &\textbf{32.60} &\textbf{35.40} \\
\bottomrule
\end{tabular}
}
\end{table}
%%%%%%%%%%%%%%%%%%%%%%%%%%%%%%%%
\begin{table}[!h]
\caption{Test accuracy on CIFAR-100 with milder data heterogeneity ($\alpha=0.5$).}
\label{tb:alpha0.5}
\vspace{-0.5mm}
\centering
\resizebox{0.55\textwidth}{!}{
\begin{tabular}{lcccccc}
\toprule
\multirow{2}{*}{Method} & \multicolumn{6}{c}{CIFAR-100}\\
\cmidrule(lr){2-7}
&Local & 1/64 & 1/16 & 1/4 & 1.0 &Union\\
\midrule
HeteroFL &30.73  &25.64  &28.36  &33.16  &35.76  &29.18     \\
FedRolex &19.06  &3.28  &10.52  &27.28  &35.16  &18.81    \\
ScaleFL  &40.64  &35.48  &43.16  &43.20  &40.72  &37.30    \\
FIARSE   &41.13  &37.88  &43.08  &43.80  &39.76  &38.91     \\
FRAMP    &\textbf{42.22}  &\textbf{40.96}  &\textbf{43.20}  &\textbf{43.88}  &\textbf{40.84}  &\textbf{40.17} \\
\bottomrule
\end{tabular}
}
\end{table}
%%%%%%%%%%%%%%%%%%%%%%%%%%%%%%%%
\begin{table}[!h]
\caption{Test accuracy on CIFAR-100 with milder data heterogeneity ($\alpha=0.7$).}
\label{tb:alpha0.7}
\vspace{-0.5mm}
\centering
\resizebox{0.55\textwidth}{!}{
\begin{tabular}{lcccccc}
\toprule
\multirow{2}{*}{Method} & \multicolumn{6}{c}{CIFAR-100}\\
\cmidrule(lr){2-7}
&Local & 1/64 & 1/16 & 1/4 & 1.0 &Union\\
\midrule
HeteroFL &31.16 &28.60 &28.80 &31.40 &35.84 &29.69     \\
FedRolex &19.06 &3.12 &11.76 &26.80 &34.56 &18.80   \\
ScaleFL  &42.05 &39.96 &43.28 &\textbf{43.44} &\textbf{41.52} & 38.44   \\
FIARSE   &40.31 &38.20 &41.44 &41.24 &40.36 &37.71    \\
FRAMP    &\textbf{43.25}  &\textbf{44.84}  &\textbf{44.32}  &42.72  &41.12  &\textbf{40.53} \\
\bottomrule
\end{tabular}
}
\end{table}
%%%%%%%%%%%%%%%%%%%%%%%%%%%%%%%%
\begin{table}[!h]
\caption{Test accuracy on CIFAR-100 with IID split.}
\label{tb:iid}
\vspace{-0.5mm}
\centering
\resizebox{0.55\textwidth}{!}{
\begin{tabular}{lcccccc}
\toprule
\multirow{2}{*}{Method} & \multicolumn{6}{c}{CIFAR-100}\\
\cmidrule(lr){2-7}
&Local & 1/64 & 1/16 & 1/4 & 1.0 &Union\\
\midrule
HeteroFL &31.87  &26.88  &31.40  &33.36  &35.84  &29.11     \\
FedRolex &20.55  &2.68  &12.24  &30.16  &37.12  &19.31    \\
ScaleFL  &44.37  &41.80  &44.76  &\textbf{46.52}  &44.40  &41.15    \\
FIARSE   &42.38  &40.12  &43.56  &44.52  &41.32  &40.63     \\
FRAMP    &\textbf{46.20}  &\textbf{45.88}  &\textbf{48.32}  &45.84  &\textbf{44.76}  &\textbf{43.06} \\
\bottomrule
\end{tabular}
}
\vspace{-1.5mm}
\end{table}

%%%%%%%%%%%%%%%%%%%%%%%%%%%%%%%%
\subsection{More Participate clients}\label{app:20paticipants}
\textbf{Participation rate of 20\%}\hspace{0.7em}
\cref{tb:participate20acc} reports test accuracy on CIFAR-100 when the client participation ratio is set to 20\%.
For FRAMP, the gains are less pronounced for clients with large submodels, but it delivers more balanced performance overall, particularly benefiting clients with limited resources, its primary design objective.

\begin{table}[h]
\caption{Test accuracy of submodels across four submodel sizes with 20\% participation rate.}
\label{tb:participate20acc}
\centering
\resizebox{0.55\textwidth}{!}{
\begin{tabular}{lcccccc}
\toprule
\multirow{2}{*}{Method} & \multicolumn{6}{c}{CIFAR-100} \\
\cmidrule(lr){2-7}
&Local & 1/64 & 1/16 & 1/4 & 1.0 &Union\\
\midrule
HeteroFL &32.23  &28.32  &31.52  &33.96  &35.12  &30.30   \\
FedRolex &33.00  &21.36  &34.12  &36.72  &39.80  &31.33   \\
ScaleFL  &39.57  &37.92  &39.60  &41.84  &38.92  &37.63   \\
FIARSE   &42.27  &40.32  &43.28  &\textbf{43.52}  &\textbf{41.96}  &38.97   \\
FRAMP    &\textbf{42.72} &\textbf{41.80} &\textbf{44.80} &43.28 &41.00 &\textbf{39.59} \\
\bottomrule
\end{tabular}
}
\end{table}

%%%%%%%%%%%%%%%%%%%%%%%%%%%%%%%
% \subsection{Sensitivity of $\lambda$}\label{app:sensityhyper}
\subsection{\texorpdfstring{Sensitivity of $\lambda$}{Sensitivity of lambda}}\label{app:sensityhyper}
To examine the sensitivity of FRAMP to the alignment loss weight $\lambda$, we conduct experiments on CIFAR-10 with $\lambda$ in ${0.1,0.2,\dots,0.9}$. For each setting, models are evaluated across submodel sizes $\gamma' \in {1/64,1/16,1/4,1.0}$, along with overall averages for Local and Union test set performance.

\begin{figure}[h] 
\centering
\includegraphics[width=0.7\textwidth]{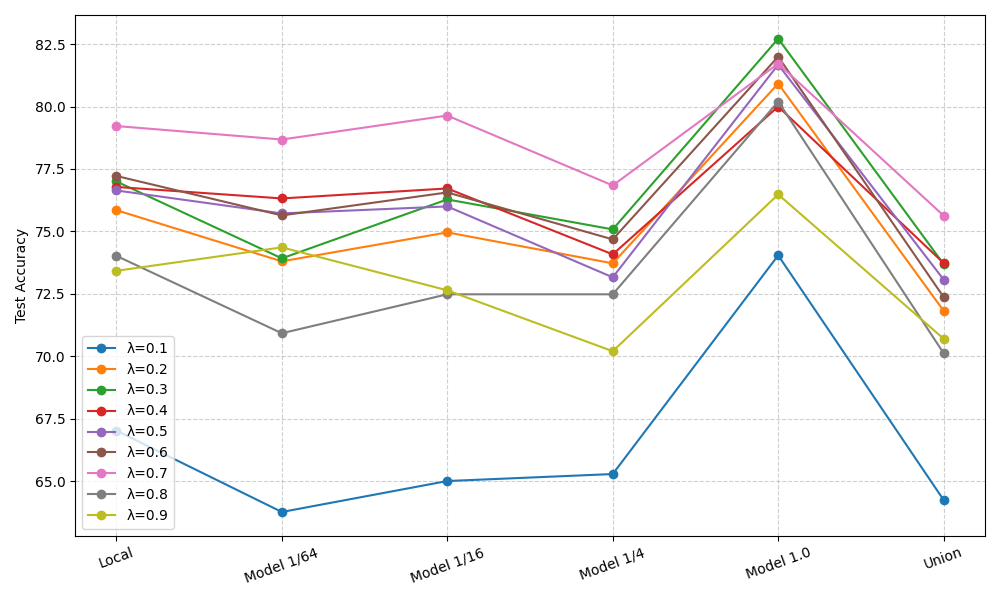} 
\caption{Test accuracy across different model sizes (Local, 1/64, 1/16, 1/4, 1.0, Union) under varying values of the alignment weight $\lambda$.}
\label{fig:lamdasens}
\end{figure}

The results in \cref{fig:lamdasens} show that FRAMP maintains stable performance for a broad range of $\lambda$. Moderate values, e.g., $\lambda=0.5$–$0.7$, yield the best overall accuracy. Very small ($\lambda=0.1$) or large ($\lambda=0.9$) values lead to reduced accuracy, reflecting under- or overemphasis on the alignment term.

%%%%%%%%%%%%%%%%%%%%%%%%%%%%%%%%
\subsection{Computational and Communication Overhead Discussion}\label{app:overhead}
In FRAMP, the additional communication overhead compared to existing heterogeneous FL baselines is:
\begin{itemize}
    \item \textbf{Client descriptor.} Each client uploads its descriptor once during initialization. A 128-dimensional descriptor is about 0.5 KB.
    \item \textbf{Class prototypes.} Class prototypes are exchanged each communication round. For ResNet18 on CIFAR-10, a full set of prototypes (10 classes × 512 dimensions) is about 20 KB.
\end{itemize}

\textbf{Scale with model size.} 
% The hypernetwork does not necessarily generate the full model as a single vector.
Model parameters can be produced in a layer-wise or block-wise manner using low-dimensional latent kernels that are reshaped, sliced, or concatenated to obtain the final parameters \cite{zhang2018graph}. This design enables the HNs to support large-scale models.

\textbf{Scale with number of clients.}
In each communication round, the server generates full models only for the participating clients. These models can be produced sequentially, keeping the peak server memory footprint effectively stable regardless of the total number of clients.
Storing client descriptors requires only lightweight data and has a negligible impact on memory consumption.

%%%%%%%%%%%%%%%%%%%%%%%%%%%%%%%%
\subsection{Privacy Discussion}\label{app:privacyanal}
Specifically, we apply two types of noise \cite{kim2025subgraph}:
1) Gaussian noise. Each prototype is perturbed as $\tilde{\bP}_n^c = \bP_n^c + \epsilon$, where $\epsilon \sim \mathcal{N}(0, \sigma^2 I)$ and $\sigma=a\cdot\|\bP_n^c\|$.
2) Random rotation. Each prototype is transformed by a random orthogonal matrix, i.e., $\tilde{\bP}_n^c = \bQ \bP_n^c$ where $\bQ\T\bQ=I$.
As shown in \cref{tb:noiseacc}, both noise injection strategies result in minimal accuracy degradation. This suggests that FRAMP leverages prototypes for semantic guidance rather than precise reconstruction and is robust to moderate obfuscation.

\begin{table}[h]
\caption{Test accuracy under prototype perturbations on CIFAR-100.}
\label{tb:noiseacc}
\centering
\resizebox{0.6\textwidth}{!}{
\begin{tabular}{lcccccc}
\toprule
\multirow{2}{*}{Method} & \multicolumn{6}{c}{CIFAR-100} \\
\cmidrule(lr){2-7}
&Local & 1/64 & 1/16 & 1/4 & 1.0 &Union\\
\midrule
NoNoise            &42.95  &43.00  &44.08  &43.72  &41.00 &40.26     \\
GN $a=0.01$        &42.40  &43.36  &44.64  &41.80  &39.80 &40.26    \\
GN $a=0.1$         &41.45  &43.20  &43.04  &41.12  &38.44 &40.04    \\
GN $a=0.5$         &41.25  &43.04  &43.32  &41.00  &37.64 &38.41    \\
Random Rotation    &41.73  &43.52  &44.24  &40.84  &38.32 &39.02  \\
\bottomrule
\end{tabular}
}
\end{table}

%%%%%%%%%%%%%%%%%%%%%%%%%%%%%%%%%%
% \subsection{Discussion of \cref{sec:existlimitation}}\label{app:furthercomparison}
\subsection{\texorpdfstring{Discussion of \cref{sec:existlimitation}}{Discussion of Section~\ref{sec:existlimitation}}}\label{app:furthercomparison}
To generate \cref{fig:lim_mask_overlap}, for each model-size group (1.0, 1/4, 1/16, 1/64), we collect the masks produced by clients in that group and compute the average activation probability for each parameter. We then plot the cumulative coverage, which shows the cumulative proportion of activated parameters covered up to a given parameter index. The curve reflects how often each parameter is selected by clients within the same model-size group.

If masks from different clients are diverse, each parameter should have a similar chance of being selected. In this case, the curve is close to a straight diagonal line, as in the size = 1.0 case. If many clients select overlapping subsets of parameters, the curve rises quickly in regions where parameters are frequently selected and becomes nearly flat in regions where parameters are rarely used. This pattern clearly appears in the size = 1/16 and 1/64 groups.

\begin{figure}[h]
\centering
\includegraphics[width=0.5\linewidth]{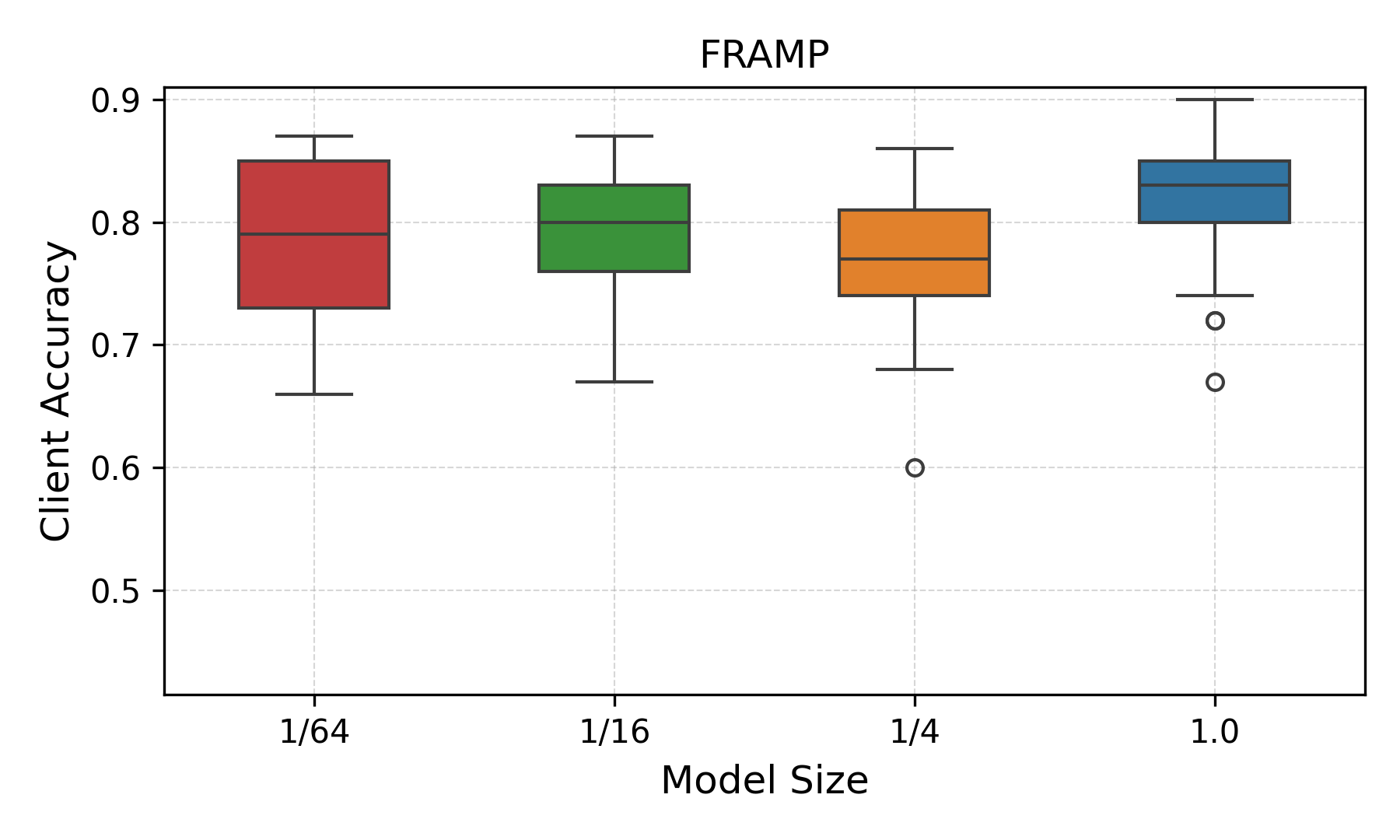}
\caption{Client accuracy distribution of submodels across different sizes. Each box represents the variation in accuracy among clients.}
\label{fig:framp_boxplot}
\end{figure}

The cumulative coverage and client accuracy distribution of FRAMP are shown in \cref{fig:ab_mask} and \cref{fig:framp_boxplot}, respectively.

%%%%%%%%%%%%%%%%%%%%%%%%%%%%%%%%%%%%%%%%%%%%%%%%%%%%%%%%%%%%%%%%%%%%%%%%%%%%%%%
%%%%%%%%%%%%%%%%%%%%%%%%%%%%%%%%%%%%%%%%%%%%%%%%%%%%%%%%%%%%%%%%%%%%%%%%%%%%%%%

\end{document}